\documentclass[10pt,twocolumn,letterpaper]{article}

\usepackage[pagenumbers]{cvpr} % To force page numbers, e.g. for an arXiv version

\definecolor{cvprblue}{rgb}{0.21,0.49,0.74}
\usepackage[pagebackref,breaklinks,colorlinks,allcolors=cvprblue]{hyperref}
\usepackage{algorithm}
\usepackage{siunitx} 
\usepackage{algpseudocode}
\usepackage{mathtools}
\usepackage{etoc}
\usepackage{soul}

\def\paperID{*****} % *** Enter the Paper ID here
\def\confName{CVPR}
\def\confYear{2026}

\title{PartiCam: Camera Controlled Video Generation with Reward Guidance}

\author{Amine Ouasfi$^{3}$,
Runjia Li$^{2}$,
Junlin Han$^{1,2}$,
Eric Marchand$^{3}$,
Philip H.S. Torr$^{2}$,
Adnane Boukhayma$^{3}$\\
$^{1}$ Meta\\
$^{2}$ University of Oxford\\
$^{3}$ INRIA, Univ. Rennes, CNRS, IRISA 
}

\begin{document}
%\maketitle

\twocolumn[{%
\renewcommand\twocolumn[1][]{#1}%
\maketitle
\begin{center}
\includegraphics[width=0.8\textwidth]{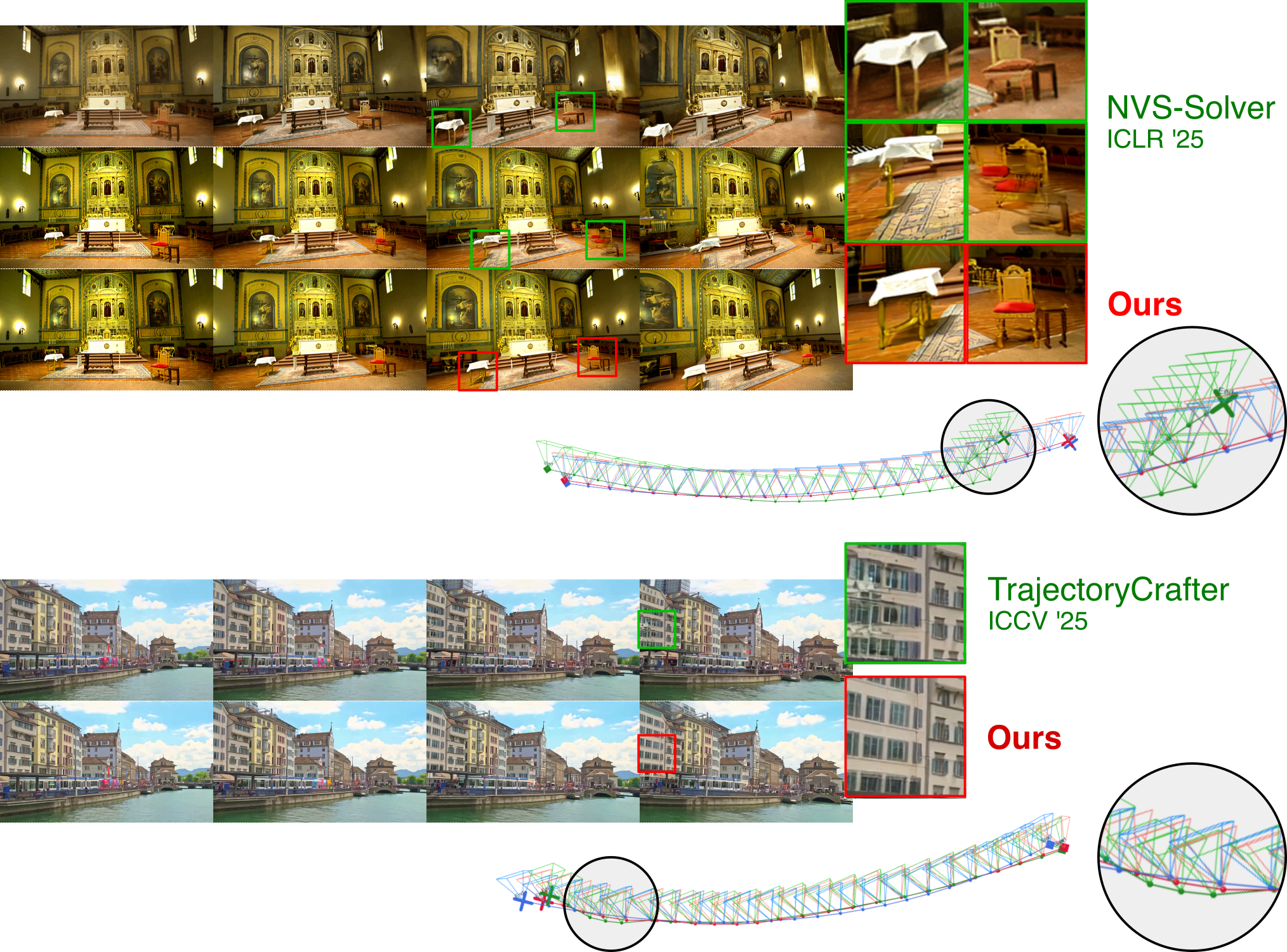}    \captionsetup{type=figure}
\caption{Our method improves camera controlled video generation. We show a top-down view of the estimated camera trajectories from the videos generated with \textcolor{red!70!black}{our method}, \textcolor{green!60!black}{its baselines} and the \textcolor{blue!70!black}{Ground-truth}. We show improvement over TrajectoryCrafter \cite{trajectorycrafter}, and NVS-Solver\cite{nvs-solver} using both SVD (top) \cite{SVD} and CogVideo (Bottom) \cite{cogvideo} backbones.}
\label{fig:teaser}
\end{center}%
}]

\section{Abstract}
We present {\bf PartiCam}, a training-free {\bf Parti}cle filtering rooted method for improved {\bf Cam}era controlled video generation. Generating videos that follow a precisely specified camera trajectory remains challenging for large video diffusion models. Training-free approaches are backbone-agnostic and avoid the need to construct large camera-annotated datasets by steering pretrained models toward the desired camera motion at test time. This enables the generation of camera-controlled video data that can subsequently be used to train camera-conditioned video diffusion models.
Existing sampling-based guidance approaches  often suffer from unstable trajectories: they either explore too broadly and fail to respect the target camera motion or collapse early and lose visual diversity over time. We introduce a global–local refinement framework for diffusion reward guidance, enabling accurate and consistent camera control during video generation. Our method builds on Sequential Monte-Carlo (SMC) guidance, but introduces a local refinement stage based on particle filtered resampling. Experiments show large improvements in camera trajectory adherence, reduced drift, and better  visual quality, without requiring model retraining. 
\section{Introduction}
\label{sec:intro}

Recent advances in multimodal generation via diffusion and flow models have
revolutionized creative fields, giving rise to video generation models
\cite{veo3,sora,moviegen,goku,malt} of impressive quality alongside emerging
zero-shot generalization capabilities \cite{wiedemer2025video}. Yet these models
often lack fine-grained control mechanisms. Building generation models that are
action-conditioned, causally and physically controllable, and 3D-grounded
represents a key step toward the vision of comprehensive world models. Open-source latent video models such as SVD \cite{SVD}, CogVideo
\cite{cogvideo,cogvideox}, Wan \cite{wan}, and Hunyuan \cite{hunyuan}, \etc provide
a practical and manageable testbed for developing ideas that can scale to larger
models and broader modalities.

Among the most important axes of control in visual data is camera viewpoint,
which accounts for a dominant source of variation in images and videos and
arguably offers the most intuitive form of control for these modalities.
The task of camera trajectory control in video generation from a single reference
image, sparse views, or a few frames of a dynamic scene is closely analogous to
few-shot novel view synthesis \cite{NeRF,3DGS}: the video model serves as a
spatio-temporal prior that enables educated completion of unobserved viewpoints.
The notorious difficulty of these tasks in classical computer vision and graphics
underscores their inherent challenges, chief among them ensuring 3D and 4D scene
consistency alongside photorealistic inpainting of occluded regions.
Camera control can be achieved by training conditional models from scratch or via adapter like fine-tuning \cite{cameractrl,trajectorycrafter}. However,
the scarcity of large-scale paired supervision, prohibitive computational training costs,
and out-of-distribution (OOD) generalization failures on challenging geometries
or novel camera trajectories can make such approaches expensive or suboptimal. A further,
more practical concern is obsolescence: open-source latent video models are evolving
rapidly, and a control module fine-tuned for one backbone is bound to that
backbone's architecture, latent space, and conditioning interface, so it risks being
superseded before it is widely adopted. Training-free methods, by contrast, are
largely backbone-agnostic and inherit the improvements of each new generation of base
models at no additional training cost.

This motivates training-free, test-time alternatives, which have become an
increasingly active direction. Beyond their immediate utility, we argue that such
methods are also a prerequisite for the training-based approaches they appear to
compete with: a reliable test-time procedure is a generator of paired
camera-annotated video data, whose samples can be  distilled back into model weights. This mirrors the trajectory of LLM
post-training, where sampling from a model and fine-tuning on the verified subset,
\ie rejection sampling fine-tuning \cite{star,rft,llama2}, converts an expensive
inference-time procedure into cheap model capability, and where the filter proved far
easier to construct than the generator.

One of the established approaches to training-free camera control is score
modulation (\eg \cite{nvs-solver}): at each denoising step, the score function is
adjusted using warped input views as scene priors, steering the latent trajectory
toward the target camera pose. We build on this formulation without loss of
generality as our underlying control mechanism. Despite its effectiveness, score
modulation operates over a \emph{single} denoising trajectory. Errors made early
in sampling, when an unlucky noise realization places the model on a poor
generative path, compound through subsequent steps and become increasingly
difficult to correct, leading to persistent camera drift even as the number of
inference steps increases.

One natural remedy is to introduce stochastic \emph{restarts} \cite{restart}
during sampling: by re-injecting forward-diffusion noise into an intermediate
latent and re-denoising, the model is given an opportunity to escape poor
generative trajectories and recover camera consistency. However, unrestricted
restarts are unstable in practice and can severely degrade generation quality by
discarding useful structure accumulated during denoising. To stabilize restarts,
we propose equipping them with \emph{reward-guided resampling}: at each restart,
we draw $N$ perturbed candidate continuations via noising-denoising, score each
against a camera alignment reward, and retain the best according to these scores.
Our formulation can support both differentiable and
non-differentiable reward functions, enabling flexible integration of arbitrary
camera consistency or visual quality metrics. This local procedure, which we term
PF-Restart, substantially improves trajectory coherence. Yet because it operates
independently for each candidate, it provides no mechanism to leverage a broader
population of trajectories for global exploration, and is therefore still
susceptible to collective drift when all candidates enter a similarly poor region
of latent space.

We address this by embedding local PF-Restart within a \emph{global} Sequential
Monte Carlo (SMC) framework. Rather than steering a single trajectory, our
method maintains a population of $K$ candidate video trajectories (\ie particles)
which are jointly propagated through the denoising process and globally reweighted
and resampled according to how well they satisfy the target camera trajectory and
desired visual properties. High-reward particles are preferentially retained and
diversified via local PF-Restart restarts, whereas low-reward ones are discarded,
giving the sampler a principled mechanism to continuously reallocate compute
toward geometrically coherent and visually appealing trajectories. This combination
of global population filtering and local reward-guided refinement empirically
mitigates early drift, avoids collapse into low-quality or geometrically
inconsistent solutions, and preserves sample diversity -- limitations that plague
both score modulation and unrestricted restart strategies individually. 
Importantly, our method is not limited to the training-free setting: it can be
applied on top of trained camera control baselines (\eg \cite{cameractrl,trajectorycrafter}),
mitigating some of their OOD generalization failures on challenging geometries or
out-of-distribution trajectories. 

We validate our approach through quantitative and qualitative evaluation on 
video reshooting from images and videos for static and dynamic scenes, 
following the experimental protocol of NVSSolver~\cite{nvs-solver}. 
Our method achieves state-of-the-art performance across these benchmarks, 
yielding superior visual results and stronger consistency with the desired 
camera trajectory, with improvements over both popular training-free~\cite{nvs-solver} 
and training-based~\cite{trajectorycrafter} methods.

In summary, our contributions are:
\begin{itemize}
    \item A training-free strategy for enhanced camera control in video model generation via reward-guidance combining global SMC trajectory exploration with local particle filtering steered restarts (PF-Restart).

    \item Analysis of reward functions that benefit the task of camera control in this context. 
    
    \item Demonstrated improvements over both training-free and training-based 
    camera control state-of-the-art methods, hedging against score modulation drift and OOD generalization limits respectively.
    
    \item State-of-the-art performance on video re-shooting benchmarks across 
    single-image and monocular video settings. 
\end{itemize}

\section{Related Work}

\paragraph{Controllable Diffusion Models.}
Denoising diffusion probabilistic models~\cite{ddpm} and efficient samplers
such as DDIM~\cite{ddim} and DPM-Solver~\cite{dpm_solver} have become the
dominant generative paradigm for images and video, with latent diffusion
models~\cite{ldm} enabling scalable high-resolution synthesis.
Training-based controlled generation paradigms including ControlNet~\cite{controlnet},
IP-Adapter~\cite{ipadapter}, T2I-Adapter~\cite{t2iadapter}, and
classifier-free guidance~\cite{cfg} require paired training data and considerable computational resources. In contrast, training-free Guidance enable control without parameter update and with fractional cost. Inference-time steering of diffusion models generally falls into three categories: noise/attention manipulation, gradient-based guidance, and particle filtering. Attention-based methods (\eg~\cite{attentionsurvey, videohandles,crossimageattention,p2p,nulltext}) are effective for image and video editing but difficult to adapt to complex geometric 3D constraints. Gradient-based methods (\eg ~\cite{pseudoinverse, diffusioninverse, freedom, universal_guidance}) utilize auxiliary energy functions to steer the denoising process.
DPS~\cite{DPS}, for instance injects differentiable likelihoods directly into the reverse process. However, computing gradients through a video diffusion model at every step is computationally expensive and prone to high variance. Our method belongs to the third category, using reward-weighted Sequential Monte Carlo (SMC) to steer the generation without requiring expensive gradient backpropagation.

\paragraph{Stochastic Sampling and Restart Mechanisms.}
Song~\etal~\cite{score_sde} establish predictor-corrector samplers under the SDE
framework, showing that mid-trajectory Langevin corrections reduce deviation from
the data manifold.
CCDF~\cite{ccdf} enforces conditioning via partial forward--reverse noise cycles,
and MCG~\cite{mcg} projects corrector steps onto the data manifold.
RePaint~\cite{repaint} alternates noising and denoising to harmonize inpainted regions with their surroundings;
Restart~\cite{restart} periodically re-injects noise to escape low-quality basins;
ZigZag~\cite{zigzag} parameterises the schedule as learnable forward--backward
passes.
All act as local annealing but lack principled weighting across candidates.
Our PF-Restart component formalises this as particle filtering with
reward-weighted resampling at each selected timestep, turning ad hoc noise
injection into a principled selection procedure.

\paragraph{Inference-Time Alignment and Sequential Monte Carlo.}
A growing body of work frames conditional diffusion as reward alignment via
Sequential Monte Carlo~(SMC)~\cite{smc_book,smcsamplers}, motivated by the
shift toward post-training optimisation~\cite{uehara_tutorial}. Tweedie's formula~\cite{tweedie} enables reward evaluation throughout the
denoising chain without model updates. TDS~\cite{tds} constructs twisted proposals targeting a
reward-weighted posterior; Dou and Song~\cite{dps_filtering} connect DPS to
exact Kalman filtering.
Kim~\etal~\cite{test_time_alignment} combine SMC with adaptive resampling to
avoid reward over-optimisation, and Kim~\etal~\cite{inference_time_scaling}
scale inference-time compute for flow models.
$\Psi$-Sampler~\cite{psi_sampler} shows that Gaussian prior initialisation wastes
particle diversity in low-reward regions, and proposes pCNL as a dimension-robust,
gradient-informed MCMC sampler to draw initial particles from the reward-aware
posterior, yielding consistent gains on layout, counting, and aesthetic tasks.
SMC has also been applied to protein design~\cite{smc_diffusion} and language
decoding~\cite{lm_smc}.
Training-based aligners such as DPOK~\cite{dpok}, DDPO~\cite{ddpo}, and
RAFT~\cite{raft} require parameter updates and are orthogonal to our approach.
Our global SMC stage builds on TDS weighting and shares $\Psi$-Sampler's
motivation of targeting reward-relevant regions early in the chain, extended to
the video latent space for camera trajectory alignment.

\paragraph{Camera-Controlled Generation.}
Stand-alone models like NeRF~\cite{NeRF} and 3DGS~\cite{3DGS} can learn scene-specific 3D representations renderable from any viewpoint given calibrated multi-view images. Dynamic versions can render dynamic scenes~\cite{def-gaussian,4d-gaussian, D-NeRF, HyperNeRF} by modeling typically a canonical 3D and deformation fields jointly. Data and regularization priors result in versions that are less likely to break under sparse input views~\cite{RegNeRF,FreeNeRF,pixelnerf,mvsnerf,sparsegs,pixelsplat,mvsplat}. However, reconstruction-based approaches are limited to observed geometry and struggle with large viewpoint extrapolation or disocclusions. Some of these limitations can be mitigated with hybrid methods distilling image or video diffusion priors into explicit 3D representations~\cite{VideoScene,Lyra,diffsplat}.  
Feed-forward multi-view diffusion models synthesize new viewpoints from one or few images~\cite{sv3d,cat3d,MVGenMaster,Elata,ViVid-1-to-3}. While requiring costly multi-view training data, many remain restricted to static or short-baseline settings with limited temporal consistency.  

\section{Method}
\label{sec:method}

We propose a probabilistic control framework for diffusion-based video generation models,
designed to steer the generation process toward a desired camera trajectory.
Our method operates in the latent space
of the diffusion model, maintaining and refining a population of stochastic trajectories.
The procedure combines a \emph{global} Sequential Monte Carlo (SMC) component 
for trajectory exploration and a \emph{local} refinement component 
that merges particle filtering with Restarts \cite{repaint, restart,zigzag}.

\subsection{Problem Setting}

Let $\mathcal{D}_\theta$ denote a video diffusion model with parameters $\theta$, 
which defines a stochastic generative process over a sequence of latent states along the sampling trajectory $\mathbf{z}_{0:T} = (\mathbf{z}_0, \ldots, \mathbf{z}_T)$:
\[
p_\theta(\mathbf{z}_{0:T}) = p(\mathbf{z}_T) \prod_{t=0}^{T-1} p_\theta(\mathbf{z}_t \mid \mathbf{z}_{t+1}),
\]
and the generated video is decoded from the latent $\mathbf{z}_{0}$ as  
$\mathbf{x}_{0} = \mathcal{G}_\theta(\mathbf{z}_{0})$.

We aim to sample trajectories that follow a target camera path 
and exhibit desired properties such as background consistency, motion smoothness, high aesthetic quality, quantified by  reward functions
\[
R(\mathbf{z}_{t}) \;=\; \sum_{i=1} \alpha_i R_i(\mathbf{x}_t),
\]
where each $R_i$ is a user-defined reward measuring alignment between the generated views and the intended camera trajectory or motion constraints, weighted by $\alpha_i$.

\subsection{Preliminaries: Score Modulation for Camera Control}

To control video diffusion models at inference time, we adopt score modulation via a latent transformation that adjusts the current latent $\mathbf{z}_{t}$ based on the input view and the target camera pose. Methods such as NVS-Solver \cite{nvs-solver} apply this principle by warping the conditioning image to the target viewpoint and injecting the warped features into the denoising update. Throughout the paper, we follow this formulation and treat the score of $\mathcal{D}_\theta$ as the modulated score. Please refer to \cite{nvs-solver} for exhaustive details.

A fundamental limitation of Score Modulation methods is that errors made early in sampling compound and become difficult to correct later, even when increasing the number of inference steps. When the early noise realization places sampling on a poor generative trajectory, the model often remains trapped there, resulting in camera drift confirmed by high ATE scores in our experiments (see Table~\ref{dynamic_res}). One way to mitigate this problem efficiently is to introduce stochastic perturbations in the form of  Restarts \cite{restart} during sampling to  recover correct camera trajectories by escaping such failure modes. Yet, in practice, unrestricted restarts are unstable and can degrade generation quality. To stabilize this mechanism, we propose to incorporate inference-time alignment \cite{uehara_tutorial} and convert the stochastic exploration into a structured filtering procedure.

We employ Sequential Monte Carlo (SMC) as an inference-time guidance mechanism. SMC naturally filters trajectory candidates by assigning each particle an importance weight derived from the reward and resampling the population so that high-reward trajectories, \ie those consistent with the target camera motion and desired visual quality, are preferentially retained. This filtering is applied globally across a population of particles throughout the diffusion process, allowing the sampler to continuously steer generation toward trajectories that remain both geometrically coherent and visually appealing. In parallel, each particle undergoes a local restart procedure: from its latent state $\mathbf{z}_t$, we draw $N_r$ perturbations by injecting forward-diffusion noise and then denoise each one back to time $t$. These proposals constitute a local candidate set on which we again apply reward-weighted resampling during the backward update. The combination of global population filtering and local restart-based refinement mitigates early drift, avoids collapse into low-quality or geometrically inconsistent trajectories, and yields temporally stable video samples that follow the prescribed motion while maintaining high aesthetic fidelity.

\begin{figure}[h!]
    \centering
    \includegraphics[width=1.0\linewidth]{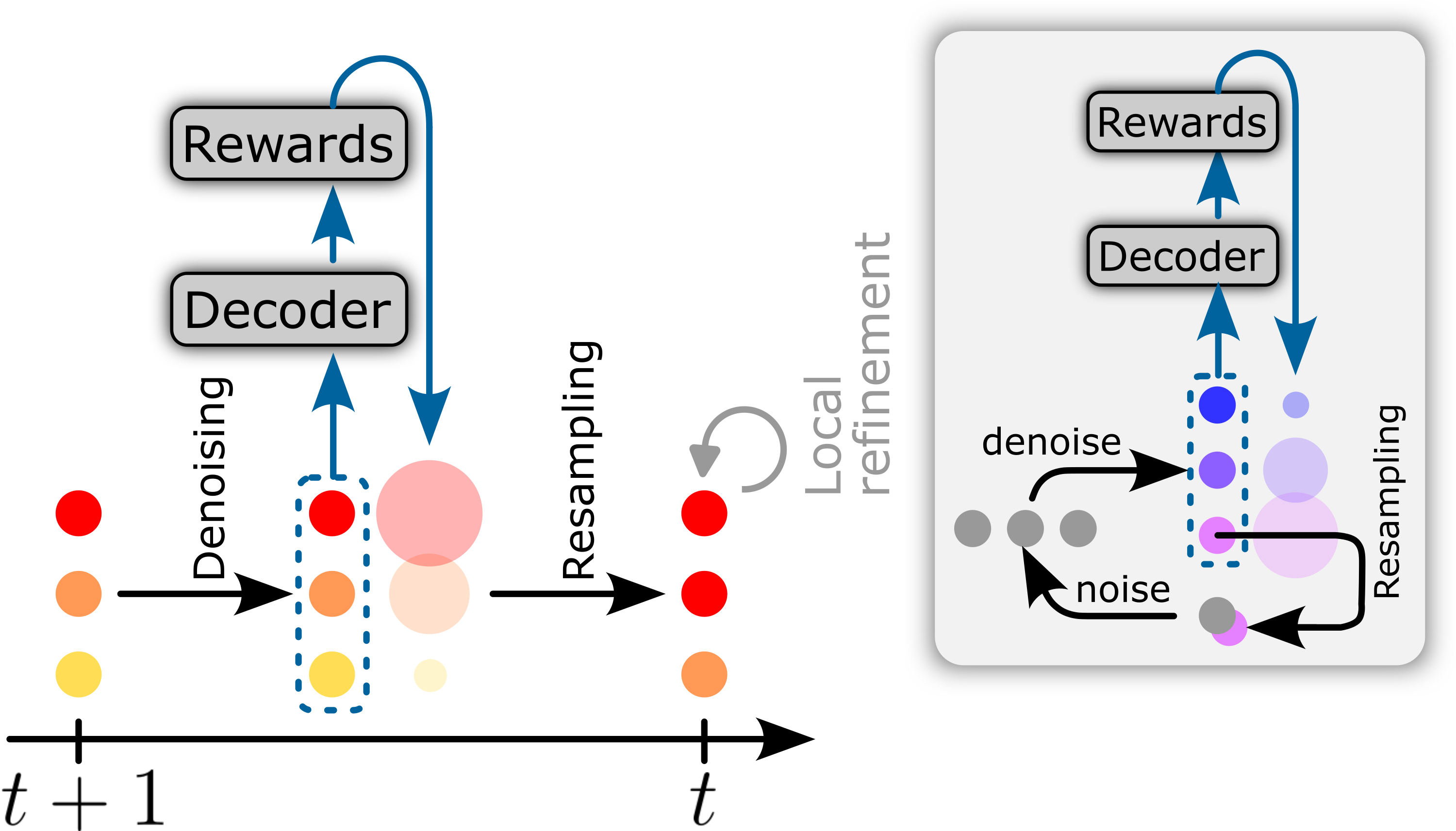}
    \caption{One denoising step with Global SMC.}
    \label{fig:global}
\end{figure}

%\begin{figure}[h!]
%    \centering
%    \includegraphics[width=0.9\linewidth]{figures/2.png}
%    \caption{On denoising step with Global SMC.}
%    \label{fig:global}
%\end{figure}

\subsection{Global Trajectory Exploration via SMC}

We aim to sample trajectories from a reward-weighted diffusion path distribution \cite{smc_book, tds}. :
\begin{equation}
p(\mathbf{z}_{0:T} \mid R)
\propto
p_\theta(\mathbf{z}_{0:T})
\exp\!\big(\beta R(\mathbf{x}_0)\big),
\end{equation}
where $\mathbf{z}_{0:T}$ denotes the latent diffusion trajectory, $p_\theta(\mathbf{z}_{0:T})$ is the diffusion prior induced by the reverse process, and $R(\mathbf{x}_0)$ is an aesthetic reward evaluated on the decoded video $\mathbf{x}_0$.

To approximate this distribution we employ a Sequential Monte Carlo sampler with $K$ particles:
\[
\{(\mathbf{z}_t^{(k)}, w_t^{(k)})\}_{k=1}^K ,
\]
which represents a weighted approximation of the intermediate target distributions.

\paragraph{Sequential target distributions.}
Following the SMC framework, we define a sequence of intermediate targets:
\begin{equation}
\gamma_t(\mathbf{z}_{t:T})
\propto
p_\theta(\mathbf{z}_{t:T})\,\psi_t(\mathbf{z}_t),
\end{equation}
where $\psi_t(\mathbf{z}_t)$ is a twisting potential representing the expected future reward conditioned on the current latent state:
\begin{equation}
\psi_t(\mathbf{z}_t)
=
\mathbb{E}\!\left[
\exp(\beta R(\mathbf{x}_0))
\,\middle|\,
\mathbf{z}_t
\right].
\end{equation}
In practice we approximate this potential by $\exp(\beta R_t(\mathbf{x}_t))$,
where $\mathbf{x}_t$ is decoded from the Tweedie estimate
$\hat{\mathbf{z}}_0(\mathbf{z}_t)$..

\paragraph{Propagation.}
At each diffusion timestep particles are propagated using the reverse diffusion transition:
\begin{equation}
\mathbf{z}_{t-1}^{(k)}
\sim
p_\theta(\mathbf{z}_{t-1} \mid \mathbf{z}_t^{(k)}),
\end{equation}
which corresponds to one denoising update of the diffusion model.

\paragraph{Weight update.}
Particle weights are updated using the incremental SMC importance weight:
\begin{equation}
\tilde w_{t-1}^{(k)}
=
w_t^{(k)}
\frac{
\psi_{t-1}(\mathbf{z}_{t-1}^{(k)})
}{
\psi_t(\mathbf{z}_t^{(k)})
}.
\end{equation}

Because we propose from the reverse transition itself,
$q_t = p_\theta(\mathbf{z}_{t-1}\mid\mathbf{z}_t)$, the prior and proposal
densities cancel in the twisted incremental weight, leaving only the ratio
of consecutive potentials. Using the practical reward approximation, this reduces to the following
\begin{equation}
\tilde w_{t-1}^{(k)}
=
w_t^{(k)}
\exp\!\Big(
\beta R_{t-1}(\mathbf{x}_{t-1}^{(k)})
-
\beta R_t(\mathbf{x}_t^{(k)})
\Big).
\label{eq:weight}
\end{equation}

Weights are normalized accordingly:
\begin{equation}
w_{t-1}^{(k)}
=
\frac{\tilde w_{t-1}^{(k)}}
{\sum_j \tilde w_{t-1}^{(j)}} .
\end{equation}

\paragraph{Resampling.}
To avoid particle degeneracy we perform resampling whenever the effective sample size is below a predefined threshold. Particles are then resampled according to their normalized weights. This global SMC filtering step is applied every $n_{smc}$ iterations to continuously bias the trajectory population toward high-reward  regions  while maintaining stochastic diversity across particles.

\subsection{Local Refinement via Guided Restarts}

To further improve local exploration around each trajectory we introduce a refinement step inspired by Restarts \cite{restart, repaint,resetting} and particle filtering.

At selected timesteps $\mathcal{T}_{\mathrm{ref}} \subseteq \{1,\dots,T\}$, each particle spawns $N_r$ candidate latents via a short noising--denoising cycle:
\begin{equation}
\tilde{\mathbf{z}}_t^{(k,n)}
=
\mathrm{Denoise}\!\big(
\mathrm{Noise}(\mathbf{z}_t^{(k)}, \sigma_{t+1})
\big),
\qquad
n = 1,\dots,N_r .
\end{equation}

This one step restart results in a diverse set of candidates that form a local proposal distribution around the current particle state.

\paragraph{Local evaluation.}
Each proposal is scored using the reward
$
\tilde w_t^{(k,n)}
\propto
\exp\!\big(
\beta R_t(\tilde{\mathbf{x}}_t^{(k,n)})
\big),
$
where $\tilde{\mathbf{x}}_t^{(k,n)}$ denotes the decoded frame corresponding to $\tilde{\mathbf{z}}_t^{(k,n)}$. The weights are normalized within each particle's proposal set
$ 
r_t^{(k,n)}
=
\frac{\tilde w_t^{(k,n)}}
{\sum_{m=1}^{N_r}} \tilde w_t^{(k,m)} .
$ 
%Note that the current particle is also included in the reward evaluation and resampling to mitigate the cases where all the candidates are worse than the current particle. 

\paragraph{Local resampling.}
A refined latent state is finally selected as follows:
\begin{equation}
\mathbf{z}_t^{(k)}
\leftarrow
\tilde{\mathbf{z}}_t^{(k,n^\ast)},
\quad
n^\ast \sim \mathrm{Cat}(r_t^{(k,1)},\dots,r_t^{(k,N_r})).
\end{equation}

This allows to locally explore the latent neighborhood while preserving high-rewards.

Since the noising--denoising cycle approximately preserves the diffusion
marginal at $t$~\cite{restart,repaint}, resampling the candidates
$\tilde{\mathbf{z}}_t^{(k,n)}$ according to $r_t^{(k,n)}$ leaves the particle
approximately distributed as $p_\theta(\cdot \mid \mathbf{z}_{t+1}^{(k)})\,
\psi_t(\cdot)$, i.e.\ the twisted proposal, at
$t \in \mathcal{T}_{\mathrm{ref}}$. Refinement thus concentrates proposals in
high-reward regions, flattening the incremental weights of \cref{eq:weight}
and reducing particle degeneracy at fixed $K$. We keep the weight update
unchanged, so that  $w_t^{(k)}$ only reflects  the selected candidate and not the
quality of the local candidate set it was drawn from~\cite{fearnhead2008particle}.
%The candidate mean of $r_t^{(k,\cdot)}$ estimates the normalizer of this proposal and could be absorbed into $w_t^{(k)}$~\cite{fearnhead2008particle}; we omit it, which biases the sampler slightly toward $p_\theta$.

For more clarity, we provide an {\bf overall algorithm description in the  supplementary material}. 

% \subsection{Overall Algorithm}

% The complete process alternates between:
% 1. global SMC propagation for exploratory trajectory sampling, and  
% 2. local PF–RePaint refinement at selected timesteps $\mathcal{T}_{\mathrm{ref}}$.

% This hybrid approach enables diffusion-based video models to 
% follow target camera trajectories without explicitly optimizing camera parameters.
% Global exploration ensures coverage of diverse motion hypotheses,
% while local refinement enforces smoothness and coherence in the generated motion.

% Algorithm \cite{alg:method} summaries our method. 

% \begin{algorithm}[h!]
% \caption{Summary of our method}
% \label{alg:method}
% \begin{algorithmic}[1]
% \Require % input(s)
% \Ensure  % output(s)

% Your steps here
%\State ...

%\end{algorithmic}
%\end{algorithm}

%\subsection{Implementation details}
\section{Implementation Details}
Our method consists of two complementary components: a local refinement module (PF-Restart) and a global exploration module based on Sequential Monte Carlo (SMC). The local refinement step corrects camera trajectories through restart operations combined with reward-guided resampling, while the global component maintains a set of reward-weighted trajectories that are periodically resampled to preserve diversity.

For local refinement, we use the \textit{camera reward}, as this stage is specifically designed to correct the camera trajectory. In contrast, the global SMC exploration relies on the \textit{aesthetic reward}. As shown in our ablations, this separation provides a favorable trade-off between aesthetic quality and camera accuracy.

In practice, we use $N_r = 2$ local candidates and $K = 2$ global particles, and run the diffusion model for $100$ iterations. To ensure reliable reward estimates, both local and global refinement are activated starting at iteration $8$. Local refinement is applied until iteration $32$, while global SMC continues until iteration $40$. Both global and local updates are performed every $4$ iterations. The local refinement step is repeated 4 times at each iteration.

These design choices allow us to maintain the overall inference time below that of \textsc{NVS-Solver} while preserving both trajectory accuracy and visual quality.
%\section{Inference time}
%Ours Method with global and local refinements takes the same time as NVS-Solver for 100 iterations (55 mins (Ours) vs 1h (NVS-Solver)). Without the global component, our method takes roughly 40 mins while significantly improving upon NVS-Solver (RTX A6000). \textbf{Note that for models other than SVD (e.g., CogVideo and Wan), \textsc{NVS-Solver} combined with DPS results in out-of-memory errors, as it requires backpropagation through the diffusion model.}
% \subsection{Overall Algorithm}

% The complete process alternates between:
% 1. global SMC propagation for exploratory trajectory sampling, and  
% 2. local PF–RePaint refinement at selected timesteps $\mathcal{T}_{\mathrm{ref}}$.

% This hybrid approach enables diffusion-based video models to 
% follow target camera trajectories without explicitly optimizing camera parameters.
% Global exploration ensures coverage of diverse motion hypotheses,
% while local refinement enforces smoothness and coherence in the generated motion.

% Algorithm \cite{alg:method} summaries our method. 

% \begin{algorithm}[h!]
% \caption{Summary of our method}
% \label{alg:method}
% \begin{algorithmic}[1]
% \Require % input(s)
% \Ensure  % output(s)

% Your steps here
%\State ...

%\end{algorithmic}
%\end{algorithm}

%\subsection{Implementation details}
\section{Experiments}
\label{sec:res}

\paragraph{Datasets} We evaluate our method on both static and dynamic scenes following the experimental setting introduced by NVS-Solver\cite{nvs-solver}. The static scenes consist of six scenes from Tanks and Temples dataset \cite{tnt} and three additional scenes selected by \cite{nvs-solver} to cover both outdoor and indoor environments. On the other hand dynamic scenes consist of nine monocular videos capturing both urban and natural settings. 
\paragraph{Metrics} We evaluate our approach in terms of camera accuracy and visual quality. For camera accuracy, we use Particle-SFM \cite{particlesfm} to estimate the camera trajectory of the generated videos and compute relative translation (RPE-T)  and rotation (RPE-R) errors \cite{rpe} as well as absolute trajectory error (ATE) \cite{rpe}. We used FID to evaluate  for visual quality of our results. 
\paragraph{Baselines} We compare to state-of-the-art methods including reconstruction ones SparseGS \cite{sparsegs}, %Sparse NeRF \cite{sparsenerf}, 
Text2NeRF \cite{text2nerf}, Def-Gaussian \cite{def-gaussian}, 4DGS \cite{4d-gaussian}, in addition to feedforward methods Photo-NVS \cite{Photoconsistent-NVS}, 3D-aware \cite{3Daware}, MotionCtrl \cite{motionctrl}, NVS-Solver \cite{nvs-solver}, TrajectoryCrafter \cite{trajectorycrafter}. Unless stated differently, our method represents our guidance applied to NVS-Solver \cite{nvs-solver} with Cogvideo \cite{cogvideo} backbone at inference time. {\bf Additional results} can be found in the {\bf supplementary material}.
\begin{figure*}[h!]
\centering
\centerline{\includegraphics[width=0.8\textwidth]{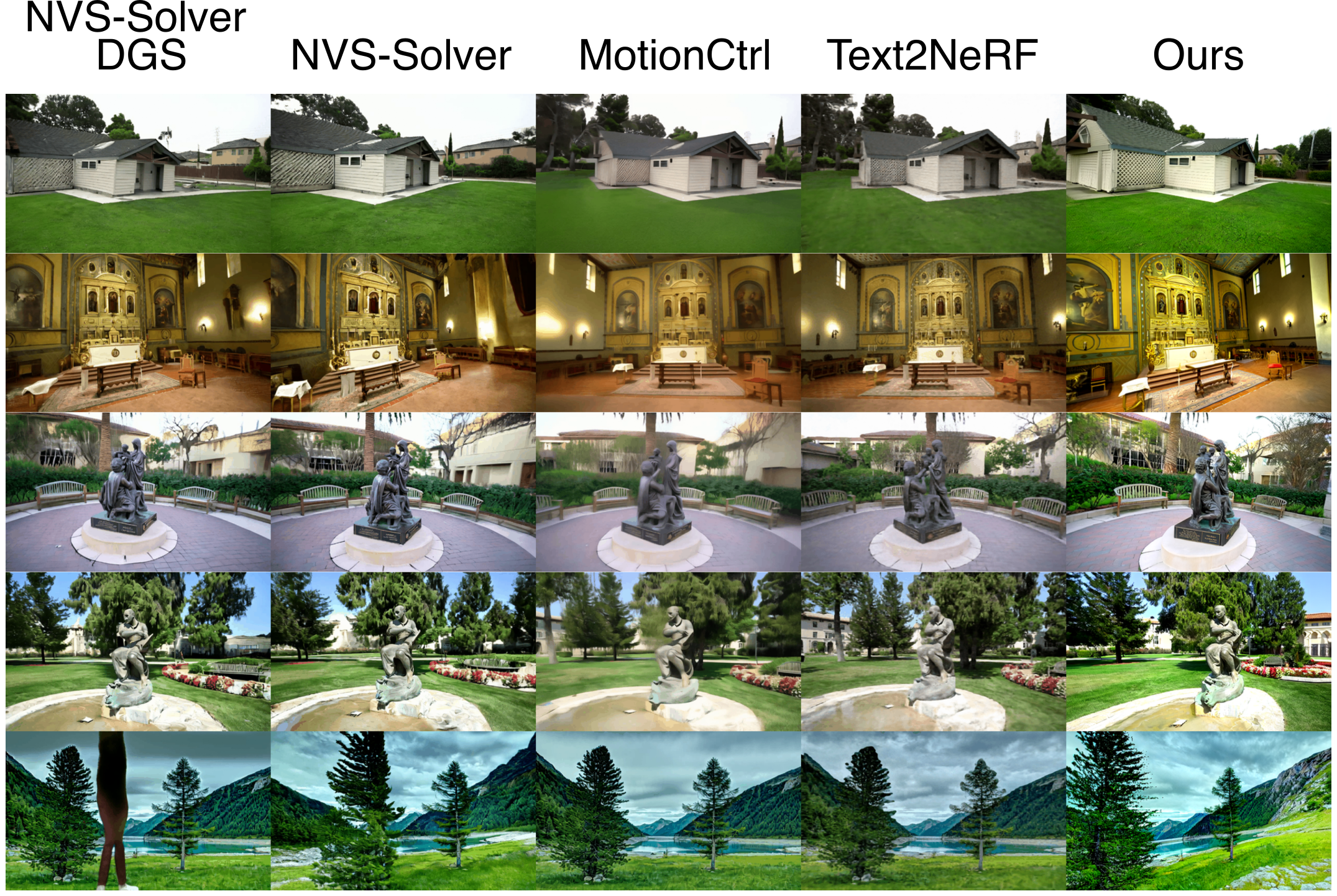}}
\caption{Qualitative comparison of different methods on NVS from single image of static scenes. We compare to methods NVS-Solver~\cite{nvs-solver}, NVS-Solver DGS~\cite{nvs-solver}, MotionCtrl~\cite{motionctrl}, Text2NeRF~\cite{text2nerf}.}
\label{fig:static_vis}
\end{figure*}

\begin{table}[h!]
  \caption{Quantitative comparison of different methods on NVS from single image of static scenes. \textit{For all metrics, the lower, the better.}}
  %\vspace{-0.3cm}
  \label{static_res}
  \centering
  \resizebox{1.0\linewidth}{!}{\setlength{\tabcolsep}{2.0mm}{
  \begin{tabular}{c|c|cccc}
    \toprule[1.2pt]
    Methods &Overfitting&  FID  &ATE &RPE-T &RPE-R   \\\hline
    SparseGS~\cite{sparsegs} &\checkmark &369.19&--&--&--  \\ 
    %Sp. NeRF~\cite{sparsenerf} &\checkmark &--&--&--&--  \\
    Text2NeRF~\cite{text2nerf} &\checkmark &187.05 &2.223&0.718& 0.107 \\
    Photo-NVS~\cite{Photoconsistent-NVS} & $\times$ &193.87 & 7.64 & 1.19 & 1.45 \\ 
    3D-aware~\cite{3Daware} &$\times$ &217.19 & 2.836&1.258 &1.662  \\ 
    MotionCtrl~\cite{motionctrl} &$\times$ &179.24&3.851&0.705&0.835   \\ 
    NVS-S {\small (DGS)} \cite{nvs-solver} &$\times$ &166.50&4.533&0.810&0.742   \\
    NVS-S {\small (Post)} \cite{nvs-solver} &$\times$ &165.12&0.767&0.156&0.170   \\\hline
    \textbf{Ours} &$\times$ &\textbf{121.56}&\textbf{0.526}&\textbf{0.122}&\textbf{0.144} \\
    \bottomrule[1.2pt]
  \end{tabular}}}
  %\vspace{-0.6cm}
\end{table}

We evaluate the camera control  in static scenes by generating novel-views from single image following prescribed camera trajectories. A quantitative comparison is presented in Table \ref{static_res}, while Figure \ref{fig:static_vis} offers a visual comparison to state-of-the-art methods. Our approach, based on reward guidance outperforms other methods in terms of camera accuracy as well as in visual quality. For camera accuracy, our method is significantly better at preserving the global trajectories, as evidenced by ATE errors compared to NVS-Solver. In terms of visual quality, our method is able to generate better structures in the occluded regions, as reflected by the FID results.  

Figure \ref{fig:teaser}  shows an additional visual comparison in this setting. We illustrate cases where our strategy can improve on top of a training-free method (NVS-Solver), as well as a state-of-the-art training-based method (TrajectoryCrafter) as backbone, both in synthesis quality and input camera trajectory adherence.

\subsection{Camera Control in Dynamic Scenes}

\begin{figure*}[!h]
\centering
\centerline{\includegraphics[width=0.8\textwidth]{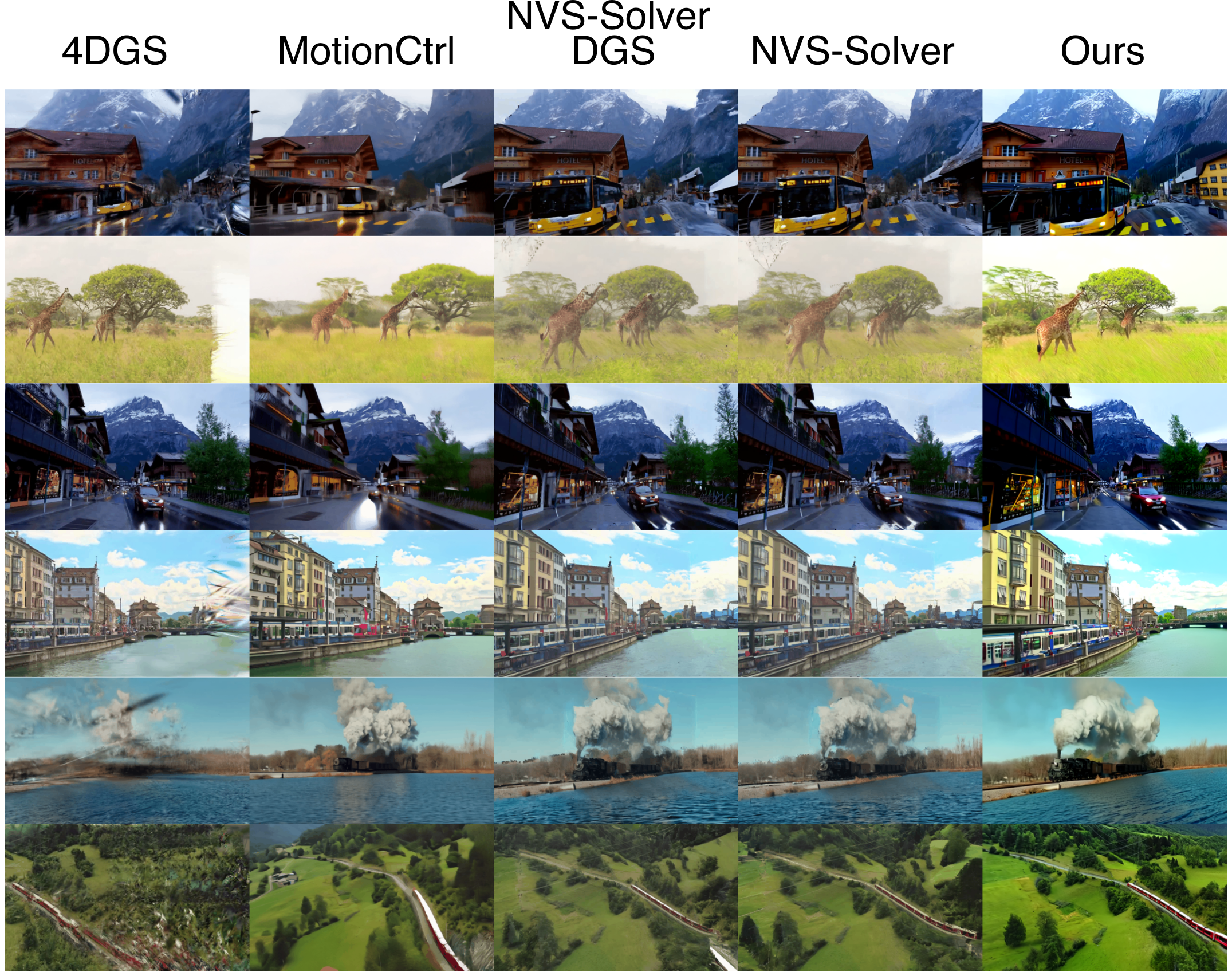}}
\caption{Qualitative comparison of different methods on NVS from monocular videos of dynamic scenes. We compare to methods 4DGS  \cite{4d-gaussian}, MotionCtrl~\cite{motionctrl}, NVS-Solver~\cite{nvs-solver}, NVS-Solver DGS~\cite{nvs-solver}.}
\label{fig:dynamic_vis}
\end{figure*}

\begin{table}[h!]
  \caption{Quantitative comparison of different methods on NVS from monocular videos of dynamic scenes. \textit{For all metrics, the lower, the better.}}
  %\vspace{-0.2cm}
  \label{dynamic_res}
  \centering
  \resizebox{1.0\linewidth}{!}{\setlength{\tabcolsep}{1.0mm}{\begin{tabular}{c|c|cccc}
    \toprule[1.2pt]
    Methods&Overfitting&FID &ATE &RPE-T &RPE-R \\\hline
    Def-Gaussian \cite{def-gaussian} &\checkmark &115.82&1.813&0.678&0.613 \\ 
    4DGS \cite{4d-gaussian} &\checkmark &74.34&2.087&0.625&0.825 \\ 
    3D-aware \cite{3Daware} &$\times$ &159.03&3.100&1.343&1.368 \\ 
    MotionCtrl \cite{motionctrl} &$\times$ &70.35&3.384&1.069&0.653 \\ 
    NVS-S {\small (DGS)} \cite{nvs-solver} &$\times$ &37.973&2.236&0.691&0.446 \\
    NVS-S {\small (Post)} \cite{nvs-solver} &$\times$ &39.86&2.308&0.725&0.400 \\\hline
    \textbf{Ours} &$\times$ &\textbf{31.86}&\textbf{0.807}&\textbf{0.061}&\textbf{0.414} \\ 
    \bottomrule[1.2pt]
  \end{tabular}}}
\end{table}

We evaluate camera control on dynamic scenes using monocular videos as input while enforcing prescribed camera trajectories. Quantitative results are reported in Table~\ref{dynamic_res}, and qualitative comparisons are shown in Figure~\ref{fig:dynamic_vis}. 

Our reward-guided sampling strategy consistently improves camera accuracy compared to existing approaches. In particular, methods such as NVS-Solver (training-free) and MotionCtrl (training-based) often struggle to follow the target trajectory in the presence of scene dynamics and occlusions. In contrast, our method achieves substantially lower Absolute Trajectory Error (ATE), as well as lower relative pose errors (RPE-T and RPE-R), indicating more accurate and faithful recovery of the desired camera motion.

Beyond trajectory accuracy, our approach also improves the visual quality of the generated videos. By exploring multiple candidate trajectories during sampling and reallocating computation toward high-reward solutions, the method better handles disocclusions and dynamic regions. As a result, the generated videos exhibit sharper structures, more consistent geometry, and fewer motion artifacts, leading to more realistic and temporally coherent renderings.

\subsection{Ablation Studies}

% \paragraph{Method ablation}
\begin{figure*}[t]
\centering

\begin{subfigure}{0.48\linewidth}
\centering
\includegraphics[width=\linewidth,trim=0 0 0 0,clip]{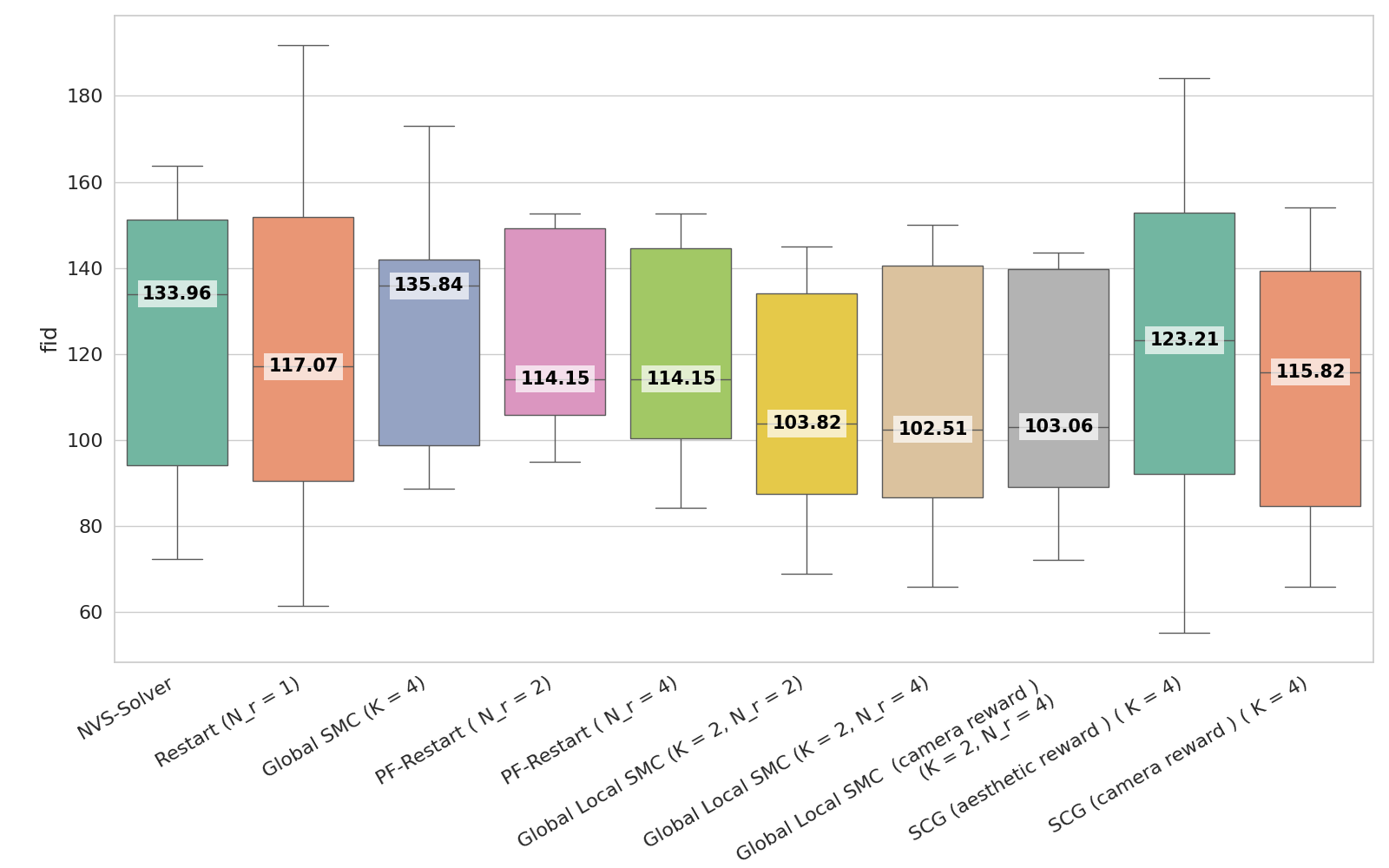}
\caption{FID}
\end{subfigure}
\hfill
\begin{subfigure}{0.48\linewidth}
\centering
\includegraphics[width=\linewidth,trim=0 0 0 0,clip]{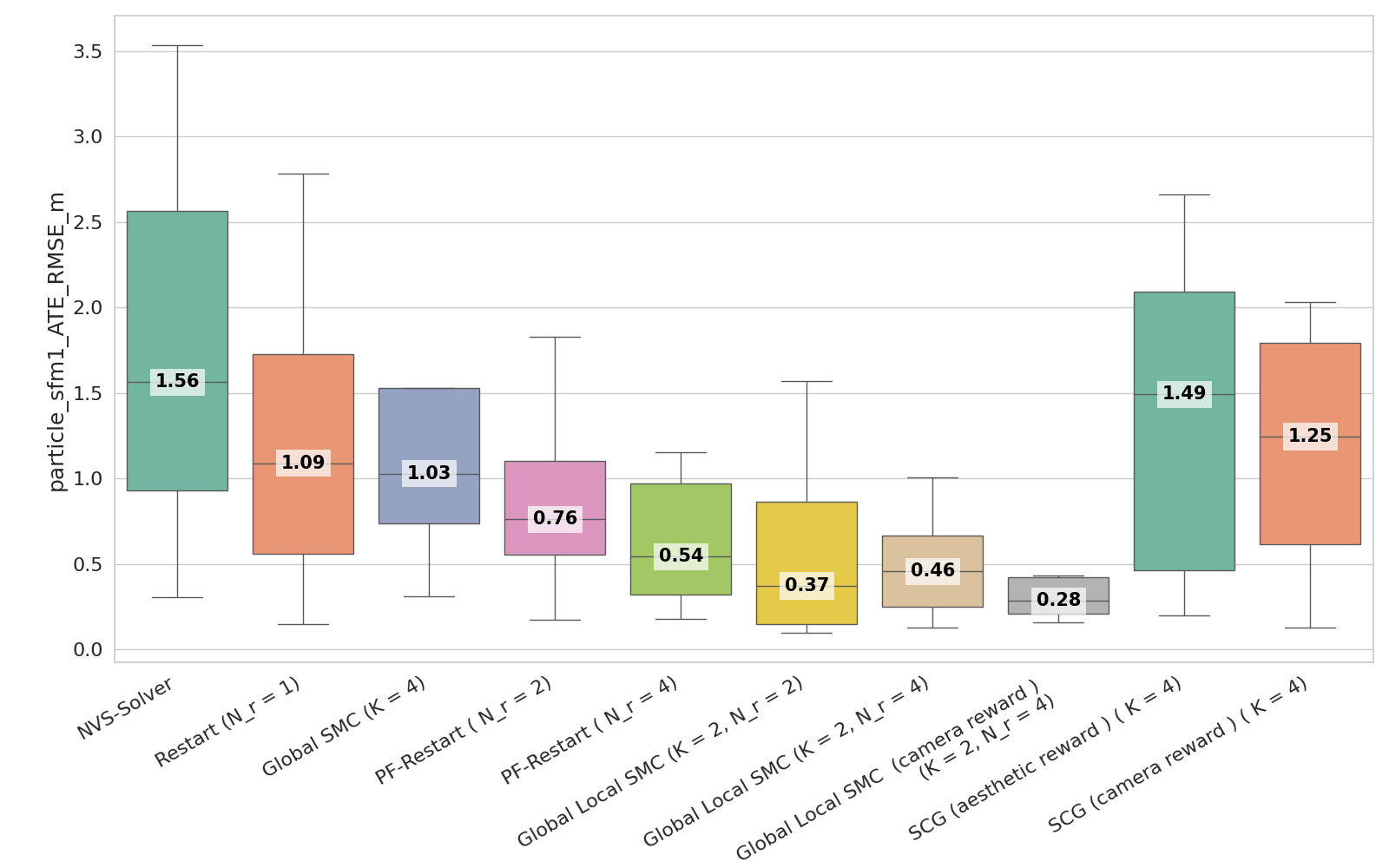}
\caption{ATE}
\end{subfigure}

\vspace{3mm}

\begin{subfigure}{0.48\linewidth}
\centering
\includegraphics[width=\linewidth,trim=0 0 0 0,clip]{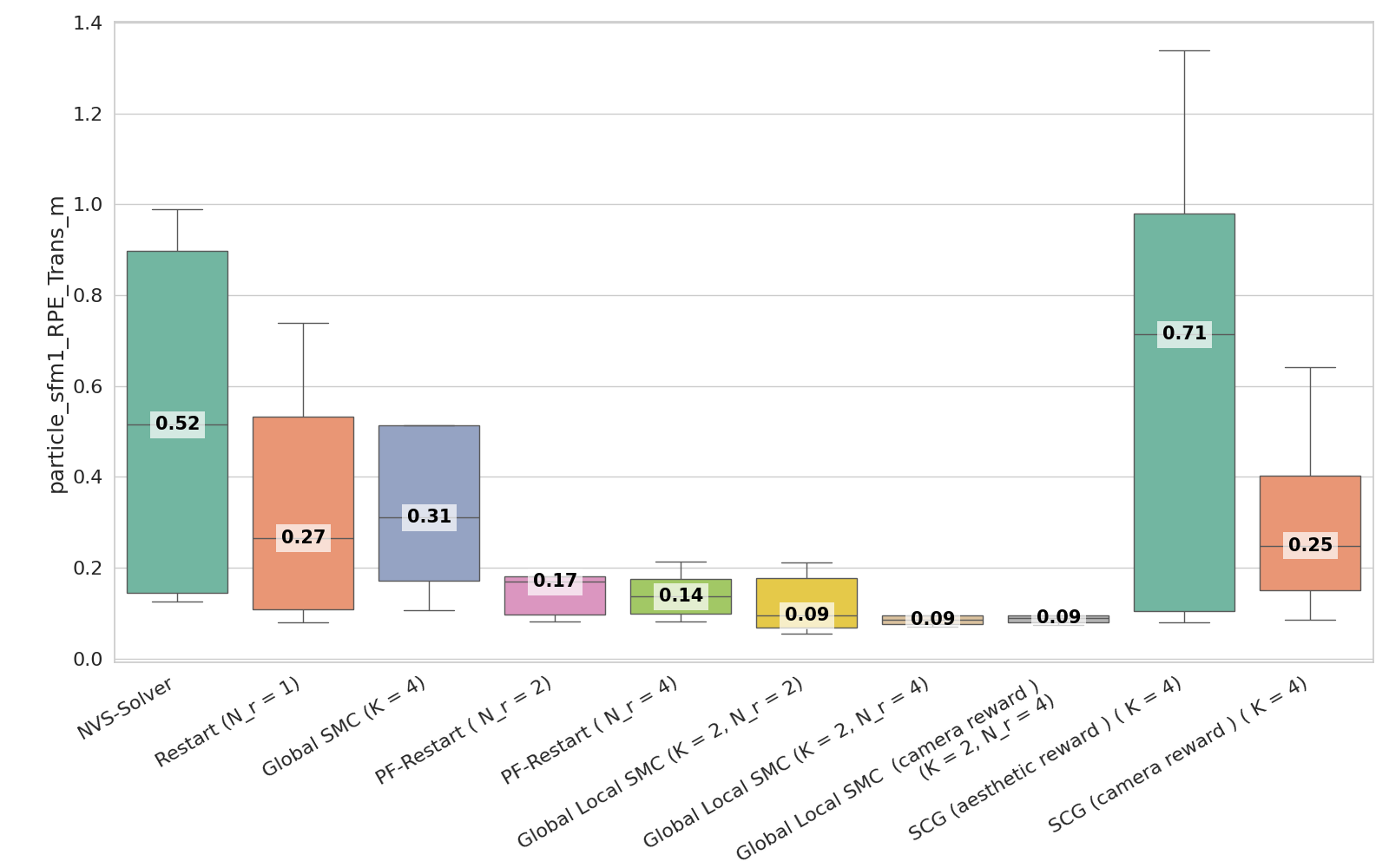}
\caption{RPE Translation}
\end{subfigure}
\hfill
\begin{subfigure}{0.48\linewidth}
\centering
\includegraphics[width=\linewidth,trim=0 0 0 0,clip]{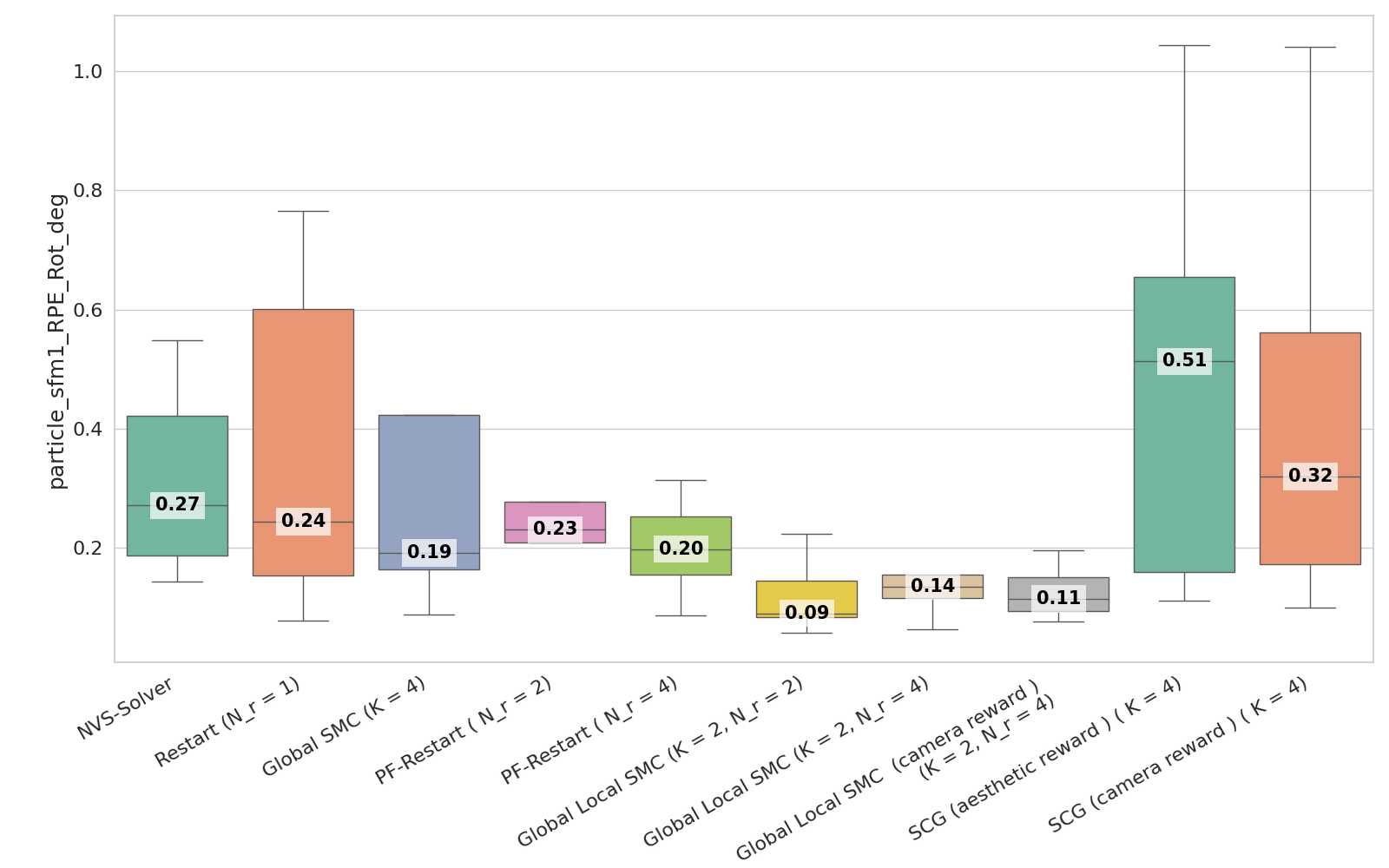}
\caption{RPE Rotation}
\end{subfigure}
\caption{Ablation results of our method on Static Scenes. We use Aesthetic reward for Global SMC and Camera reward for local refinement. The number of local candidates is denoted as $N_r$ while the number of global particles is referred to as $K$.   }
\label{fig:ablation}
\end{figure*}

\begin{table}[h!]
  \caption{Ablation study of our method with respect to the choice of the backbone video model for NVS from monocular videos of dynamic scenes. \textit{For all metrics, the lower, the better.}}
  %\vspace{-0.2cm}
  \label{model_ablation}
  \centering
  \resizebox{1.0\linewidth}{!}{\setlength{\tabcolsep}{1.0mm}{\begin{tabular}{c|cccc}
    \toprule[1.2pt]
    Methods&FID &ATE &RPE-T &RPE-R \\\hline
    NVS-S (CogVideo) \cite{nvs-solver} &41.08& 0.981 &0.128&0.431 \\ 
    Ours (CogVideo)  &31.86& 0.807 & 0.061 & 0.414\\  
    NVS-S (SVD) \cite{nvs-solver}  &39.86&2.308&0.725&0.400 \\ 
    Ours (SVD)  &38.26&1.912&0.521&0.489 \\ 
    TrajectoryCrafter \cite{trajectorycrafter}   &29.86&0.767&0.095&0.460\\ 
    Ours (TrajectoryCrafter)  &30.01&0.712&0.070&0.428 \\ 
    \bottomrule[1.2pt]
  \end{tabular}}}
\end{table}

We conduct an ablative analysis on the static scenes from the NVS-Solver benchmark~\cite{nvs-solver} using right-sweep trajectories, evaluating each component of our method using ATE, RPE-T, RPE-R, and FID. Results are summarised in Figure~\ref{fig:ablation}.
\paragraph{Effect of restart sampling} Starting from the NVS-Solver baseline (ATE: 1.56, RPE-T: 0.52, RPE-R: 0.27, FID: 133.96), introducing a simple restart without reward guidance already yields meaningful improvements across all metrics (ATE: 1.09, RPE-T: 0.27, RPE-R: 0.24, FID: 117.07). However, as evidenced by the wide interquartile ranges in Figure~\ref{fig:ablation}, plain restart suffers from high variance across scenes:  while some scenes benefit substantially, others show little or no improvement. This indicates that without any reward signal to guide resampling, noise injection makes the generation process unstable.

\paragraph{Effect of reward-guided local refinement (PF-Restart)} Incorporating reward-guided resampling into the restart procedure  (PF-Restart)  significantly tightens this variance while further improving trajectory accuracy. With $N_r = 2$ local candidates, PF-Restart reduces ATE to 0.76   and RPE-T to 0.17 . Increasing to $N_r = 4$  pushes these to 0.54   and 0.14   respectively, while maintaining FID at 114.15 in both settings. Compared to unguided restart, the reward-guided resampling yields a notably more compact distribution of per-scene errors, confirming that reward alignment stabilizes the local search.
\paragraph{Effect of global SMC exploration} Global Sequential Monte Carlo sampling alone (K=4) reduces ATE to 1.03 and RPE-R to 0.19, but fails to achieve satisfactory performance on translation accuracy (RPE-T: 0.31 ) and actually degrades FID relative to NVS-Solver (135.84 vs.\ 133.96). This counterintuitive finding highlights a fundamental limitation of population-level exploration without local correction: while global diversity prevents early collapse into poor basins, it does not provide the fine-grained per-particle guidance needed to correct individual trajectory errors. This result confirms that global SMC alone is insufficient.
\paragraph{Complementarity of global and local components} Combining global SMC with local PF-Restart, our full method achieves the best overall performance and validates the complementarity of the two mechanisms. With $K=2$ and $N_r=2$, using the aesthetic reward for global exploration and the camera reward for local refinement, the joint method reaches RPE-T: 0.09 , RPE-R: 0.09, ATE: 0.37, and FID: 103.82, which is a 6× reduction in translation error and 4× reduction in ATE relative to NVS-Solver, while simultaneously improving visual quality. Increasing $N_r$ to 4 does not monotonically help: ATE slightly degrades from 0.37  to 0.46  and RPE-R worsens from 0.09° to 0.14°, suggesting that  local restarting with more particle budget can undermine global trajectory consistency when the control signal is not strong enough. The sweet spot in our setting is $K=2$,$N_r=2$.  
\paragraph{Effect of reward design} A crucial choice for our method is the choice of the 
reward function we use to guide the sampling. We experimented in the paper with camera error using VGGT \cite{vggt} to estimate trajectories in addition to aesthetic reward used in \cite{VBench}. In our default configuration, the global stage uses the aesthetic reward to preserve visual diversity across particles, while the local stage uses the camera reward to correct geometric drift. We additionally evaluate a variant where the camera reward drives both stages (
$K=2$,$N_r=4$). Replacing the aesthetic reward with the camera reward in the global SMC yields the best absolute ATE of 0.28: a further 24\% improvement over the aesthetic-global variant at the same $N_r$ and also improves RPE-R from 0.14 to 0.11 relative to the aesthetic variant. This comes at a minor cost in visual quality (FID: 103.06 vs.\ 102.51) and a slight regression in RPE-R relative to the $K=2$,$N_r=2$ default (0.11 vs.\ 0.09). These results reveal a clear trade-off: using the camera reward globally sharpens trajectory adherence but reduces the diversity of particle proposals. The default separation with aesthetic reward globally and  camera reward locally provides a good  balance between geometric accuracy and visual quality, and is the configuration adopted throughout our main experiments. Similarly, we can see that for SCG using the camera reward significantly improves the camera metrics (ATE: 1.49 vs 1.25). For PF-Restart, we noted that they both perform equally well for scenes without large occlusion as in this case the quality of the video is enough to guide the model to generate meaningful content in the occluded regions. However when the motion is ambiguous due to large occlusions, Camera reward shows significantly better results improving the Absolute Trajectory Error (ATE) from 1.00 to 0.71.
\paragraph{Comparison to other possible guidance methods} These methods differ in how they estimate reward, how they sample candidate trajectories, and how they select the best samples. We seek a method that can quickly explore the space of diffusion trajectories and accurately select the best ones. TDS~\cite{tds} and $\Psi$-Sampler~\cite{psi_sampler} are the closest to our method but they require differentiating through the reward function and to backpropagate through the diffusion model which is impractical for reward functions like the camera reward and  recent video diffusion models like Cog-Video or Wan. On the other hand, SCG \cite{beamsearch} has the advantage of maintaining one global particle and sampling candidates using the reverse SDE which results in less diverse samples throughout the iterations. With aesthetic reward it performs the worst of all evaluated configurations on trajectory metrics (RPE-T: 0.71, ATE: 1.49), confirming that a non-geometric reward is insufficient to guide camera-accurate generation under this sampler. SCG with camera reward recovers partially (RPE-T: 0.25, ATE: 1.25) but still falls far behind our full method. Consequently, beyond the choice of the reward,  our approach lies in the bi-level SMC framework itself: global exploration and local reward-guided refinement together achieve what neither component nor SCG can accomplish alone.

\paragraph{Backbone}

Our proposed  method can be applied to any latent video diffusion model. Throughout the paper, we primarily used CogVideo \cite{cogvideo}. To evaluate the generality of our approach, we perform an ablation study by applying our method to SVD \cite{SVD} and TrajectoryCrafter \cite{trajectorycrafter}.

As shown in Table \ref{model_ablation}, our method consistently improves upon NVS-Solver both in the cases where the video model backbone is SVD and CogVideo. This result highlights the effectiveness of our Camera Control strategy in mitigating the weaknesses of score modulation.

On the other hand, TrajectoryCrafter, which is a finetuned version of CogVideo specifically designed for camera control, already achieves very competitive performance even without additional guidance. Interestingly, when applying our method to TrajectoryCrafter, we observe further improvements in camera accuracy while maintaining comparable visual quality.

\subsection{Mitigating limitations of TrajectoryCrafter\cite{trajectorycrafter}}
Training based camera controlled video generation methods can display visual artefacts, inconsistencies and misalignment with the input camera trajectory, which can be framed as generalization issues. We show here examples where our method can help recover from such failures. We challenge the state-of-the-art training based method TrajectoryCrafter\cite{trajectorycrafter} with harder camera trajectories at test time, following the camera trajectory sampling in \cite{nvs-solver}. Figure \ref{fig:traj} shows our inference-time improvement over this method using  some of the demo videos of TrajectoryCrafter\cite{trajectorycrafter}. 
%Here the camera reward guidance plays a major role in this improvement.  

\subsection{Mitigating limitations of ReCamMaster\cite{recammaster}}

\begin{figure}[h!]
\centering
%\vspace{-25pt}
\includegraphics[width=1.0\linewidth]{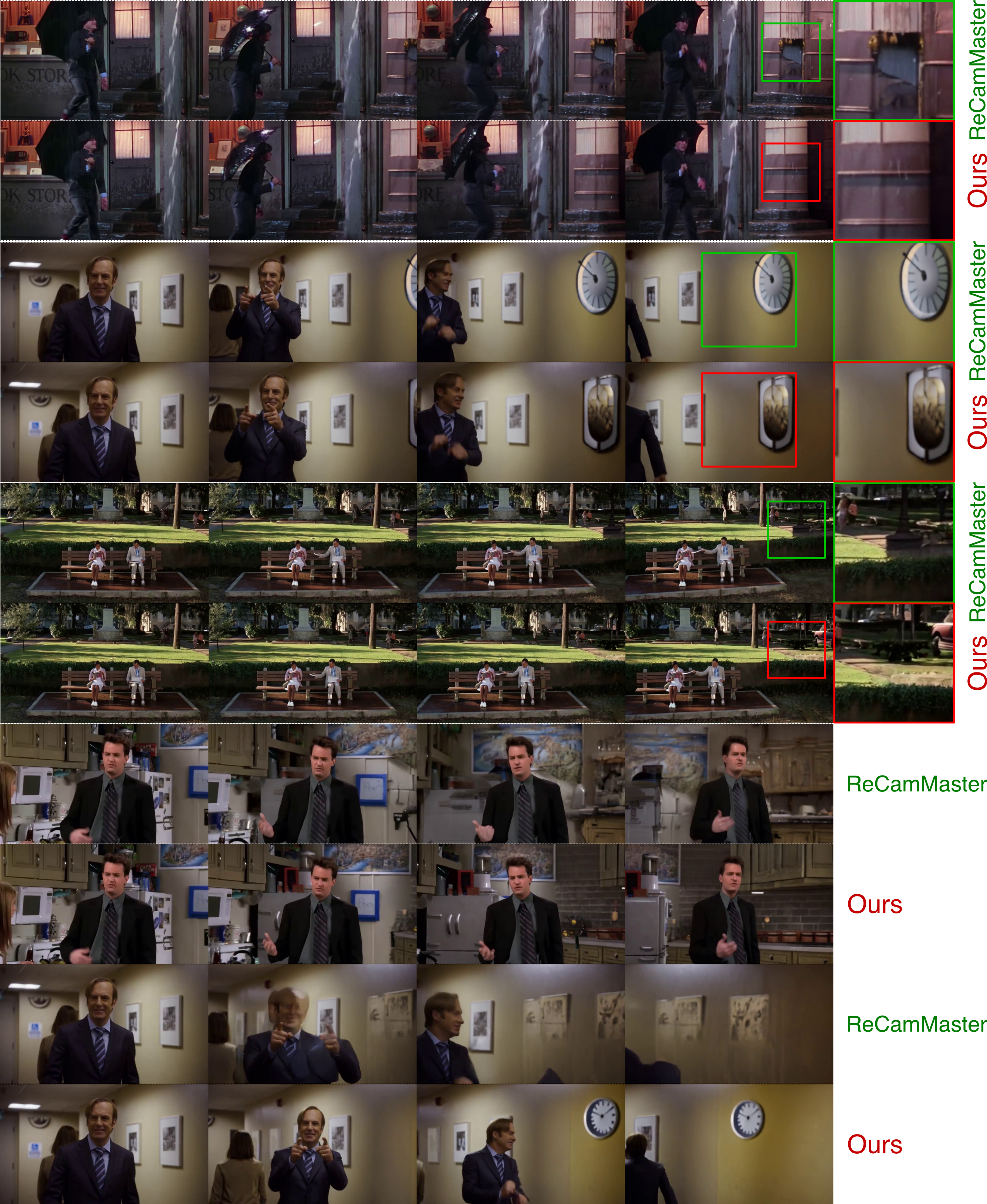}
\caption{Qualitative comparison in monocular video reshooting of dynamic scenes to ReCamMaster \cite{recammaster}.}
\label{fig:recam}
%\vspace{-25pt}
\end{figure}

\begin{figure}[h!]
\centering
%\vspace{-25pt}
\includegraphics[width=1.0\linewidth]%{prefcamtrol/figures/traj_hard}
{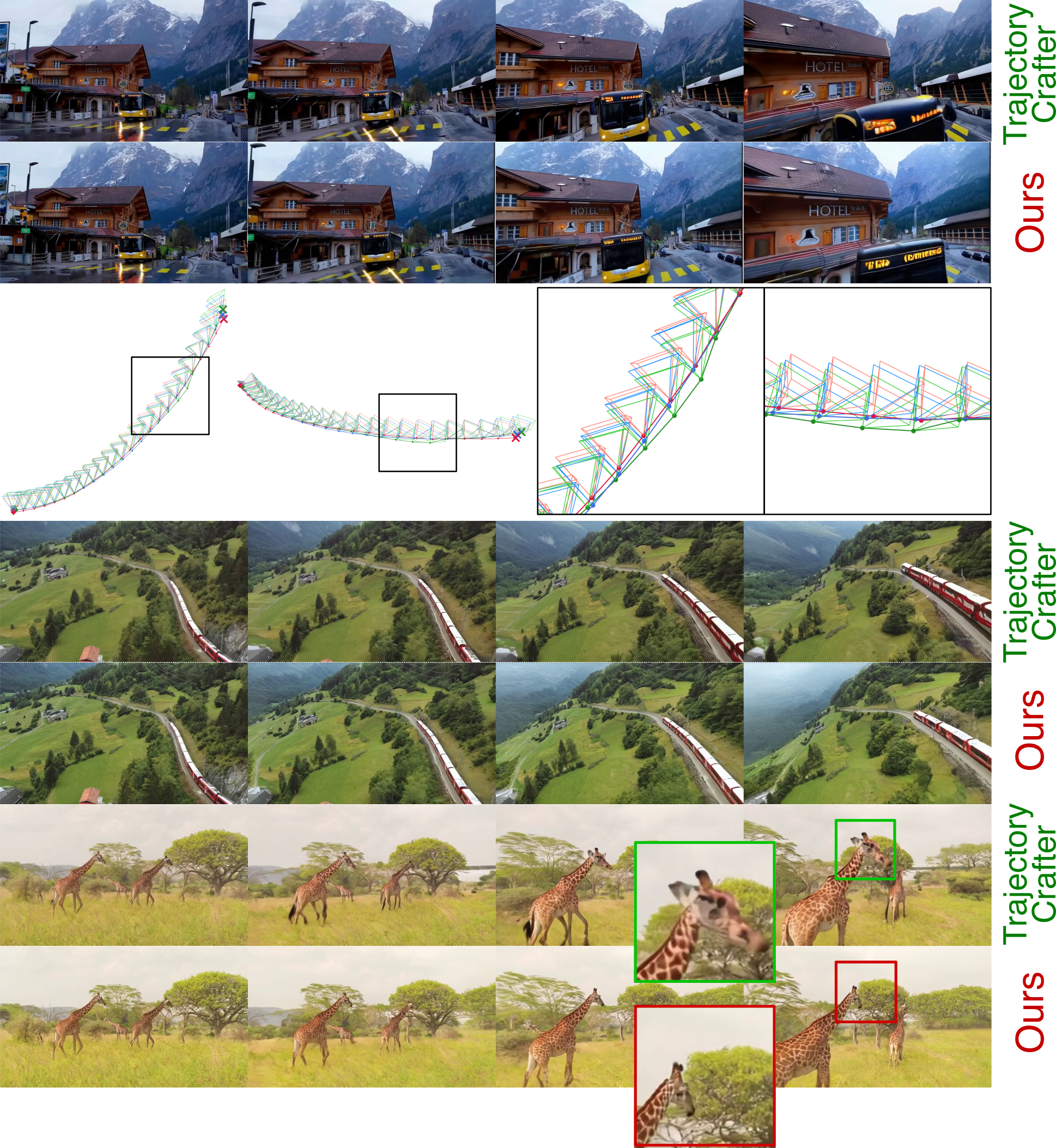}
\caption{Qualitative comparison in monocular video reshooting of dynamic scenes to TrajectoryCrafter \cite{trajectorycrafter}. We We show a two views of the estimated camera trajectories from the videos generated with \textcolor{red!70!black}{our method}, \textcolor{green!60!black}{TrajectoryCrafter} and the \textcolor{blue!70!black}{Ground-truth}.}
\label{fig:traj}
%\vspace{-25pt}
\end{figure}

Training based method ReCamMaster\cite{recammaster} displays visual artifacts especially under challenging camera trajectories. Figure \ref{fig:recam} shows our inference-time improvement over this method.

\section{Limitations}
\label{sec:limit}

Our method can require a larger denoising budget due to maintaining multiple candidate trajectories. Conversely, reward guidance can also accelerate convergence by steering sampling toward better solutions early. Overall, the approach represents a reasonable trade-off when generation quality and adherence to the control signal are prioritized. Additionally, the effectiveness of the framework depends on the design of the reward function, which may require task-specific tuning.
\section{Conclusion}
We presented PartiCam, a training-free framework for improving camera trajectory control in diffusion-based video generation. Our approach combines global Sequential Monte Carlo (SMC) trajectory exploration with local particle-filtered
restarts (PF--Restart), enabling reward-guided sampling that balances exploration and refinement during the denoising process. This formulation mitigates common failure modes of existing approaches, including early trajectory drift in score-modulated sampling and limited robustness to OOD camera motions. By maintaining a population of candidate trajectories and reallocating sampling effort toward high-reward
solutions, our method produces videos that better adhere to the desired camera trajectory while maintaining visual quality and temporal coherence.

Beyond camera control, the proposed framework provides a general mechanism for reward-guided steering of diffusion processes without requiring gradient backpropagation or model retraining. We believe this population-based inference
strategy may extend naturally to other controlled generation tasks and forms a promising direction for future work.

{
    \small
    \bibliographystyle{ieeenat_fullname}
    \bibliography{main}
}

\section{Supplementary Material}
\section{Algorithm}

Algorithm \ref{alg:alg} summarizes our proposed method.   

\begin{algorithm}[t]
\caption{SMC Diffusion with Local Guided Restarts}
\label{alg:smc_diffusion}
\begin{algorithmic}[1]
\Require diffusion model $p_\theta$, reward $R$, particles $K$, local proposals $N_{\mathrm{r}}$, refinement steps $\mathcal{T}_{\mathrm{ref}}$, $N_{pf}$ repetitions of local refinement.
\State Initialize $\mathbf{z}_T^{(k)}\sim\mathcal{N}(0,I)$, $w_T^{(k)}=1/K$
\For{$t=T,\dots,1$}
    
    \If{$t\in\mathcal{T}_{\mathrm{ref}}$}
        \For{$k=1,\dots,N_{pf}$}
        \For{$k=1,\dots,K$}
            \State Generate $\tilde{\mathbf{z}}_t^{(k,n)}=\mathrm{Denoise}(\mathrm{Noise}(\mathbf{z}_t^{(k)},\sigma_{t+1}))$, $n=1\dots N_{\mathrm{r}}$
            \State Score $r_t^{(k,n)}\propto\exp(\beta R_t(\tilde{\mathbf{x}}_t^{(k,n)}))$ 
            \State Sample $\mathbf{z}_t^{(k)}\leftarrow\tilde{\mathbf{z}}_t^{(k,n^\ast)}$, $n^\ast\sim\mathrm{Cat}(r_t^{(k,:)})$
        \EndFor
        \EndFor
    \EndIf
    \For{$k=1,\dots,K$}
        \State Sample $\mathbf{z}_{t-1}^{(k)}\sim p_\theta(\mathbf{z}_{t-1}\mid\mathbf{z}_t^{(k)})$
        \State Update $\tilde w_{t-1}^{(k)} \leftarrow w_t^{(k)} \exp(\beta R_{t-1}(\mathbf{x}_{t-1}^{(k)})-\beta R_t(\mathbf{x}_t^{(k)}))$
    \EndFor
    \State Normalize weights $w_{t-1}^{(k)}\propto\tilde w_{t-1}^{(k)}$; resample if ESS below threshold
\EndFor
\State \Return $\{\mathbf{x}_0^{(k)}\}_{k=1}^K$
\end{algorithmic}
\label{alg:alg}
\end{algorithm}
\section{Qualitative results}

\begin{figure}[h!]
\centering
%\vspace{-25pt}
\includegraphics[width=1.0\linewidth]{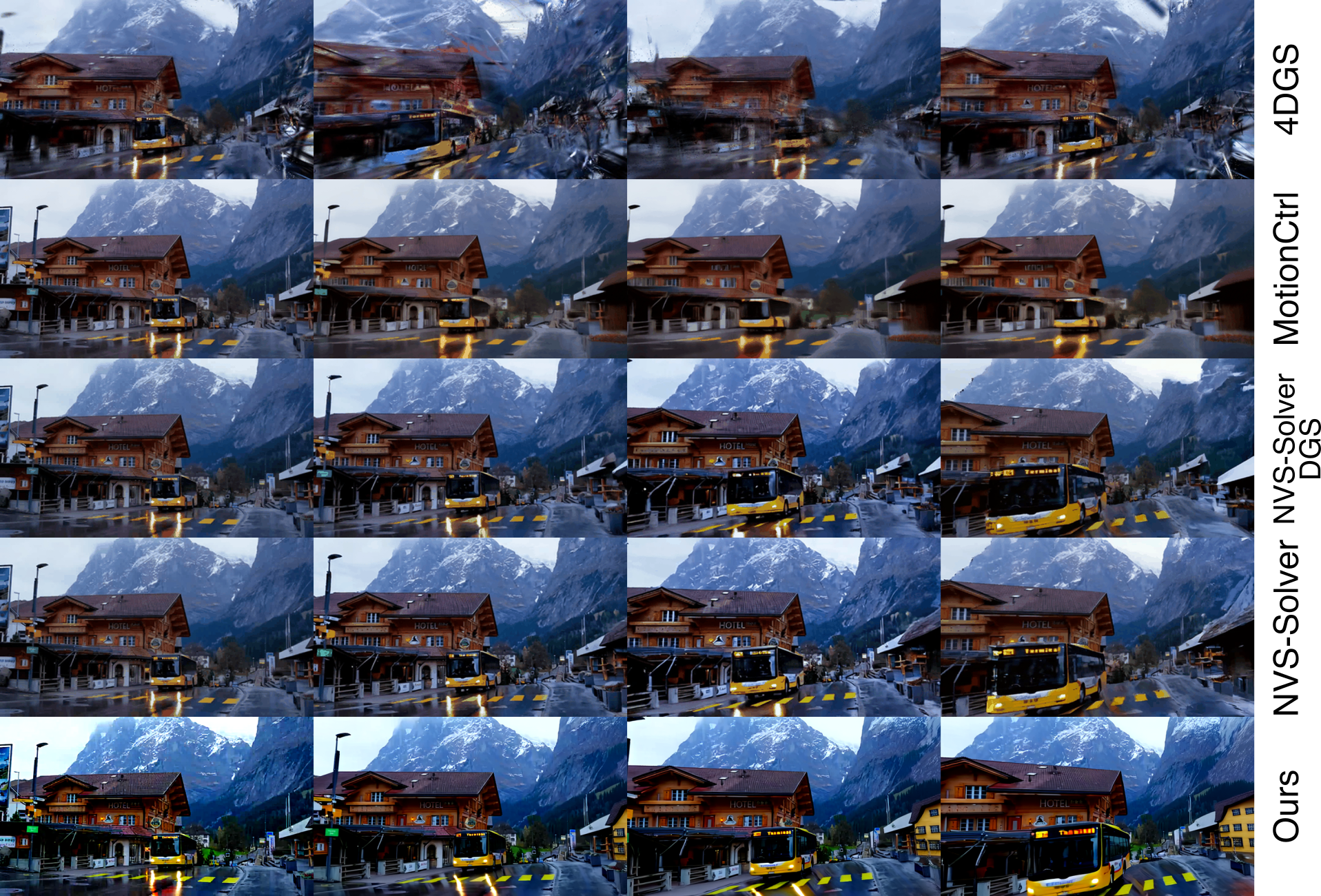}
\includegraphics[width=1.0\linewidth]{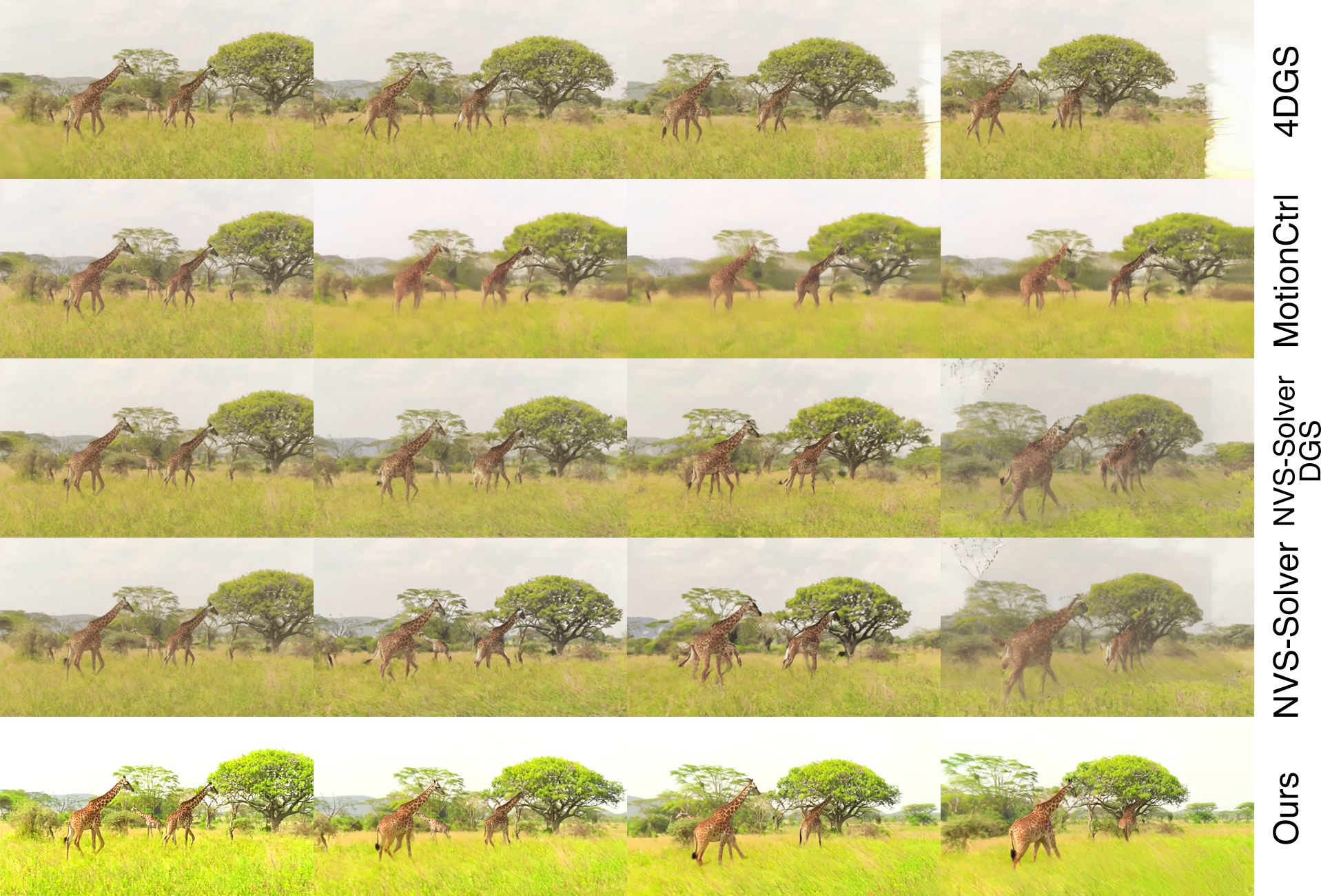}
\includegraphics[width=1.0\linewidth]{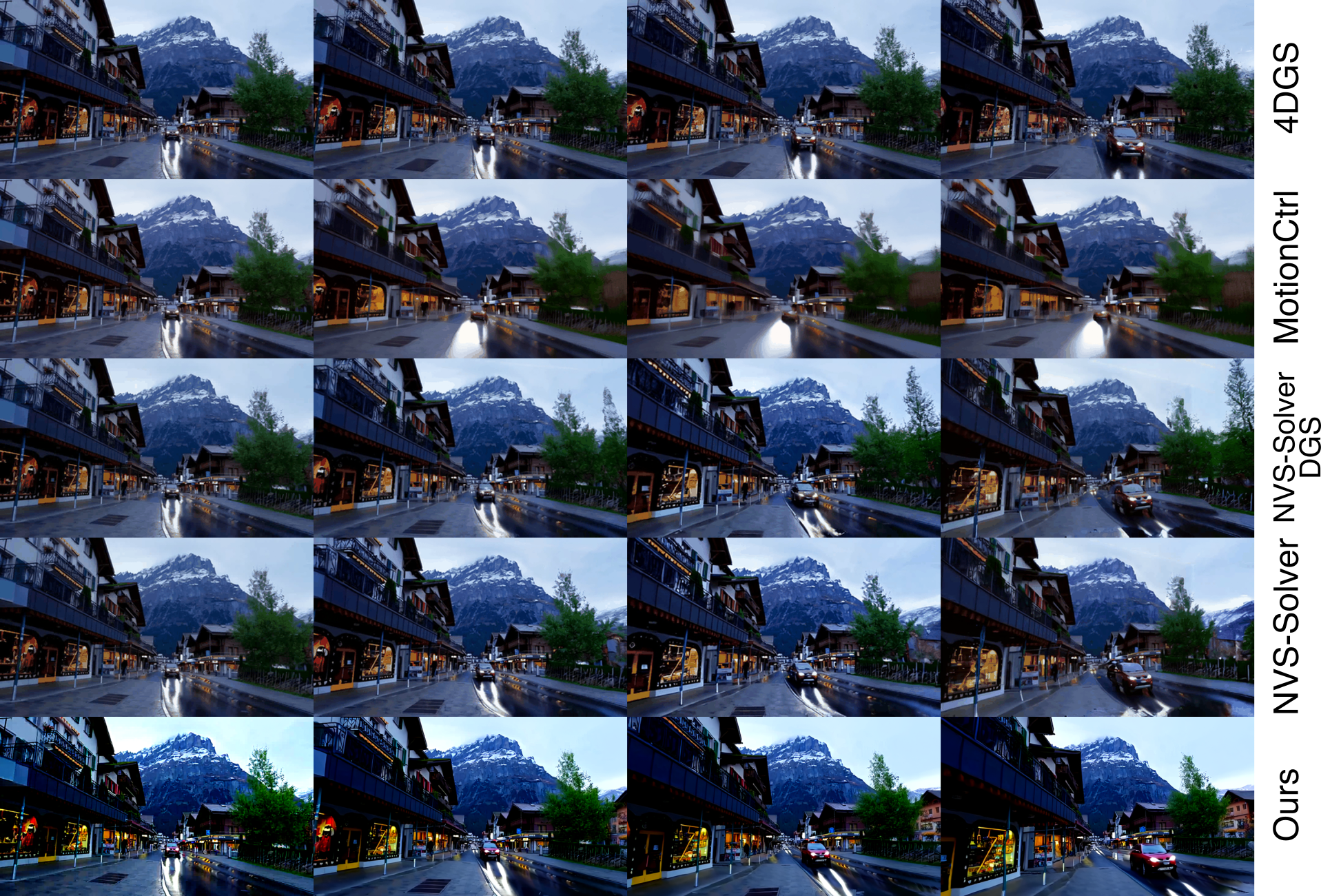}
\caption{Qualitative comparison of different methods on monocular video reshooting of dynamic scenes. We compare to methods 4DGS  \cite{4d-gaussian}, MotionCtrl~\cite{motionctrl}, NVS-Solver~\cite{nvs-solver}, NVS-Solver DGS~\cite{nvs-solver}.}
\label{fig:dyn1}
%\vspace{-25pt}
\end{figure}

\begin{figure}[h!]
\centering
%\vspace{-25pt}
\includegraphics[width=1.0\linewidth]{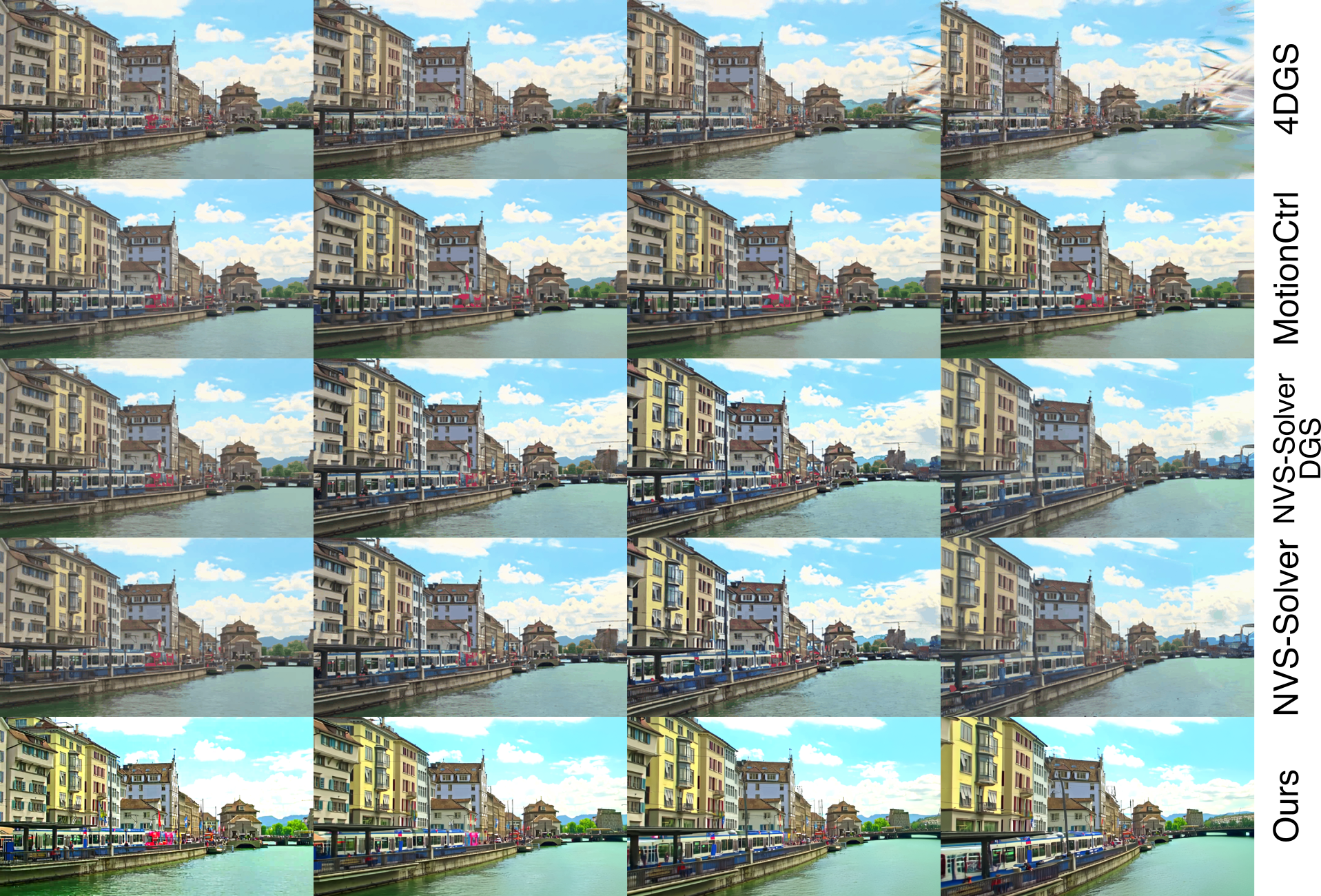}
\includegraphics[width=1.0\linewidth]{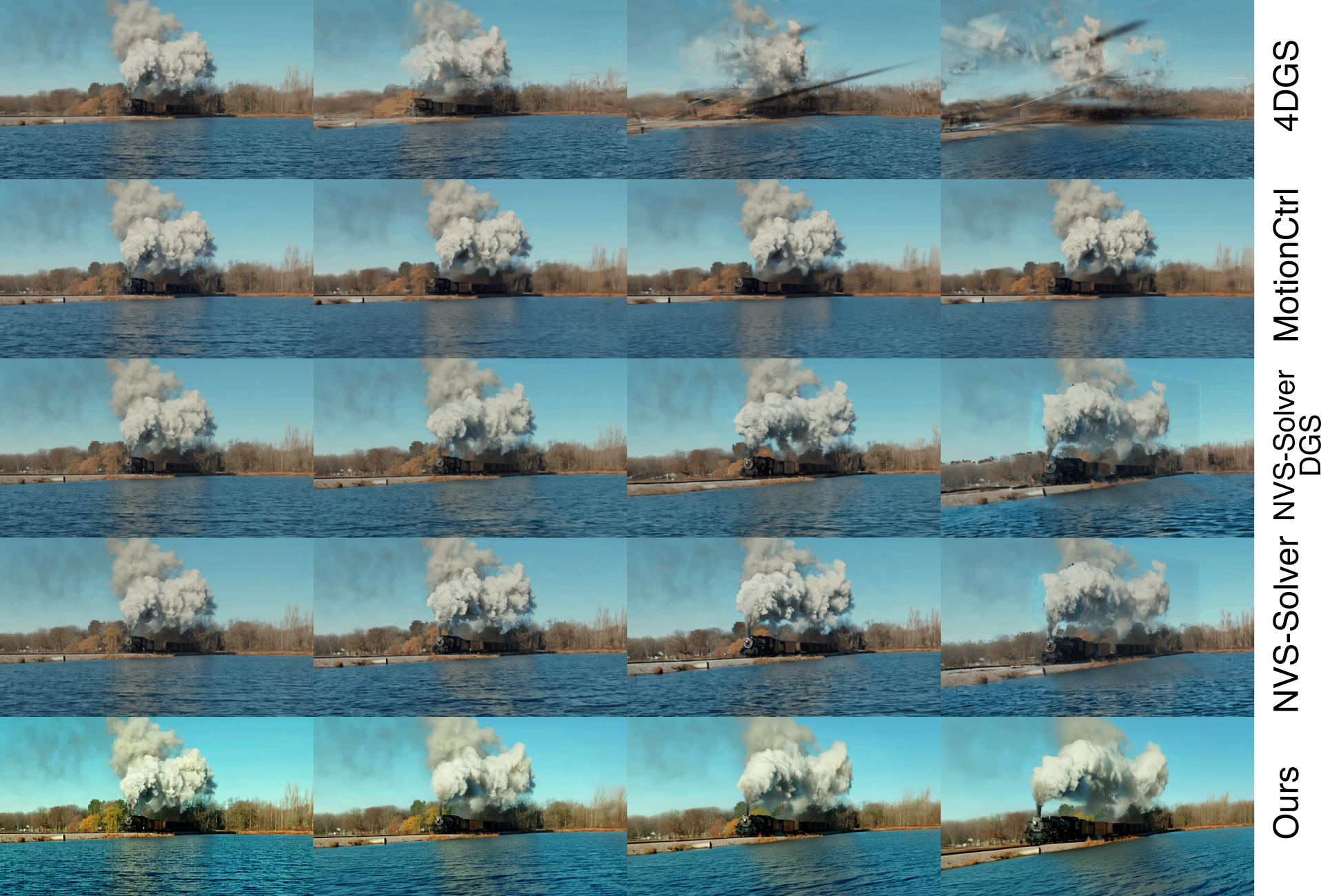}
\includegraphics[width=1.0\linewidth]{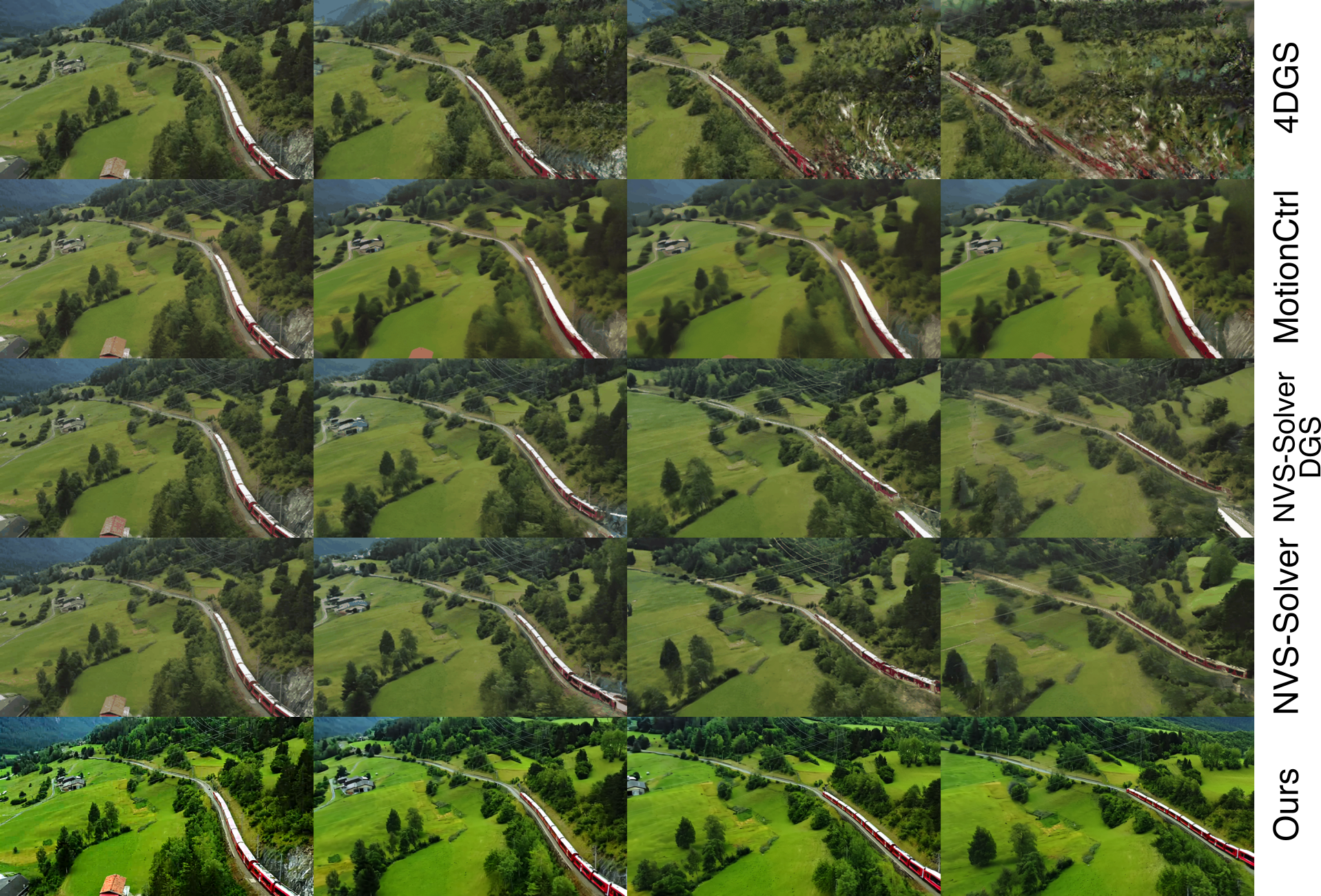}
\caption{Qualitative comparison of different methods on monocular video re-shooting of dynamic scenes. We compare to methods 4DGS  \cite{4d-gaussian}, MotionCtrl~\cite{motionctrl}, NVS-Solver~\cite{nvs-solver}, NVS-Solver DGS~\cite{nvs-solver}.}
\label{fig:dyn2}
%\vspace{-25pt}
\end{figure}

We provide additional qualitative results and comparisons in this section. Figures \ref{fig:dyn1} and \ref{fig:dyn2} complement Figure 4 in the main paper. While the main paper shows only a single-frame comparison with our method building on CogVideo \cite{cogvideo} based NVS-Solver \cite{nvs-solver}, here we present additional frames for each method.

% WARNING: do not forget to delete the supplementary pages from your submission 
% \input{sec/X_suppl}

\end{document}